%% file: main.tex
\documentclass{article} 
\usepackage{iclr2025_conference,times}

\input{math_commands.tex}

\input{commons}
\usepackage{hyperref}
\usepackage{url}

\title{\makebox[0.96\linewidth][s]{ReMoE: Fully Differentiable Mixture-of-}\\Experts ~with ~ReLU ~Routing}

\author{Ziteng Wang,~~Jun Zhu,~~Jianfei Chen\thanks{Corresponding author} \\
Dept. of Comp. Sci. and Tech., Institute for AI, BNRist Center, THBI Lab, \\
Tsinghua-Bosch Joint ML Center, 
Tsinghua University \\
{\fontsize{9.7}{13}\texttt{wangzite23@mails.tsinghua.edu.cn};~~\texttt{\{dcszj,jianfeic\}@tsinghua.edu.cn}}
}

\iclrfinalcopy 
\begin{document}

\maketitle

\vspace{-0.35cm}
\input{src/abstract}

\input{src/sec01-intro}
\input{src/sec02-preliminaries}

\input{src/sec03-method}

\input{src/sec04-exp}

\input{src/sec05-discussion}
\input{src/sec06-related}
\input{src/sec07-conclusion}

\bibliography{iclr2025_conference}
\bibliographystyle{iclr2025_conference}

\newpage
\appendix
\input{src/appendix}

\end{document}

%% file: math_commands.tex
\usepackage{amsmath,amsfonts,bm}

\def\eqref#1{equation~\ref{#1}}

\def\1{\bm{1}}

\def\vx{{\bm{x}}}
\def\vy{{\bm{y}}}

\def\mM{{\bm{M}}}

\def\mW{{\bm{W}}}

\DeclareMathAlphabet{\mathsfit}{\encodingdefault}{\sfdefault}{m}{sl}
\SetMathAlphabet{\mathsfit}{bold}{\encodingdefault}{\sfdefault}{bx}{n}

\newcommand{\Ls}{\mathcal{L}}
\newcommand{\R}{\mathbb{R}}



%% file: commons.tex
\usepackage{color}
\usepackage{bm}
\usepackage{hyperref}
\usepackage{amsmath,amsfonts,amsthm}
\usepackage{blindtext}
\usepackage{listings}
\usepackage{algorithm}
\usepackage{algorithmic}
\usepackage{booktabs}
\usepackage{multirow}
\usepackage{multicol}
\usepackage{caption}
\usepackage{subcaption}
\usepackage{graphicx}

\usepackage{wrapfig}

\usepackage{threeparttablex}
\usepackage{tabularx}

\usepackage{ifthen}

\newif\ifdebug
\debugtrue

\newcommand{\rebuttal}[1]{{{#1}}}

\newcommand{\norm}[1]{\left\lVert#1\right\rVert}

%% file: src/abstract.tex
\begin{abstract}
Sparsely activated Mixture-of-Experts (MoE) models are widely adopted to scale up model capacity without increasing the computation budget. However, vanilla TopK routers are trained in a discontinuous, non-differentiable way, limiting their performance and scalability. 
To address this issue, we propose ReMoE, a fully differentiable MoE architecture that offers a simple yet effective drop-in replacement for the conventional TopK+Softmax routing, utilizing ReLU as the router instead. 
We further propose methods to regulate the router's sparsity while balancing the load among experts.
ReMoE's continuous nature enables efficient dynamic allocation of computation across tokens and layers, while also exhibiting domain specialization.
Our experiments demonstrate that ReMoE consistently outperforms vanilla TopK-routed MoE across various model sizes, expert counts, and levels of granularity. Furthermore, ReMoE exhibits superior scalability with respect to the number of experts, surpassing traditional MoE architectures. The implementation based on Megatron-LM is available at \href{https://github.com/thu-ml/ReMoE}{https://github.com/thu-ml/ReMoE}.
\end{abstract}

%% file: src/sec01-intro.tex
\section{Introduction}
Transformer models
~\citep{vaswani2017attention} consistently improve performance as the number of parameters increases~\citep{kaplan2020scaling}. However, scaling these models is constrained by computation resources. Sparsely activated Mixture-of-Experts (MoE)~\citep{shazeer2017outrageously} mitigates this challenge by employing a sparse architecture that selectively activates a subset of parameters during both training and inference. This conditional computation allows MoE models to expand model capacity without increasing computational costs, offering a more efficient alternative to dense models.

The key component in MoE is the routing network, which selects the experts to activate for each token. Various routing methods~\citep{shazeer2017outrageously,lewis2021base,roller2021hash,zhou2022mixture} have been proposed, with TopK routing~\citep{shazeer2017outrageously} being the most commonly adopted. However,  the vanilla TopK router introduces a discrete and non-differentiable training objective~\citep{shazeer2017outrageously,zoph2022st}, limiting the performance and scalability.

Recent works on fully-differentiable MoE aim to overcome this limitation. Soft MoE~\citep{puigcerver2023sparse} introduces token merging, while SMEAR~\citep{muqeeth2023soft} proposes expert merging. However, both approaches break token causality, making them unsuitable for autoregressive models. Lory~\citep{zhong2024lory} improves upon SMEAR and is applicable to autoregressive models. But it underperforms vanilla MoE with TopK routing.

In this work, we address the discontinuities by introducing ReMoE, an MoE architecture that incorporates ReLU routing as a simple yet effective drop-in replacement for TopK routing. Unlike TopK routing, which computes a softmax distribution over the experts and calculates a weighted sum of the largest $K$ experts, ReLU routing directly controls the active state of each expert through a ReLU gate. The number of active experts is determined by the sparsity of the ReLU function. To maintain the desired sparsity, we propose adding a load-balancing refined $L_1$ regularization to the router outputs, with an adaptively tuned coefficient. This approach ensures that ReMoE maintains the same computational costs as TopK-routed MoE.

Compared to TopK routing, ReLU routing is continuous and fully differentiable, as the ReLU function can smoothly transition between zero and non-zero values, indicating inactive and active. Besides, ReLU routing manages the ``on/off'' state of each expert independently, offering greater flexibility. Moreover, the number of activated experts can vary across tokens and layers, enabling a more efficient allocation of computational resources. Further analysis reveals that ReMoE effectively learns to allocate experts based on token frequency and exhibits stronger domain specialization.

Our experiments on mainstream LLaMA~\citep{touvron2023llama} architecture demonstrate that ReLU routing outperforms existing routing methods including TopK routing and fully-differentiable Lory. Through an extensive investigation across model structures, we find that ReMoE consistently outperforms TopK-routed MoE across a broad range of active model sizes (182M to 978M), expert counts (4 to 128), and levels of granularity (1 to 64)~\citep{krajewski2024scaling}. Notably, in terms of scaling behavior, we observe that ReMoE exhibits a steeper performance improvement as the number of experts scales up, surpassing traditional MoE models.

%% file: src/sec02-preliminaries.tex
\begin{figure}[t]
    \centering
    \includegraphics[width=1\linewidth]{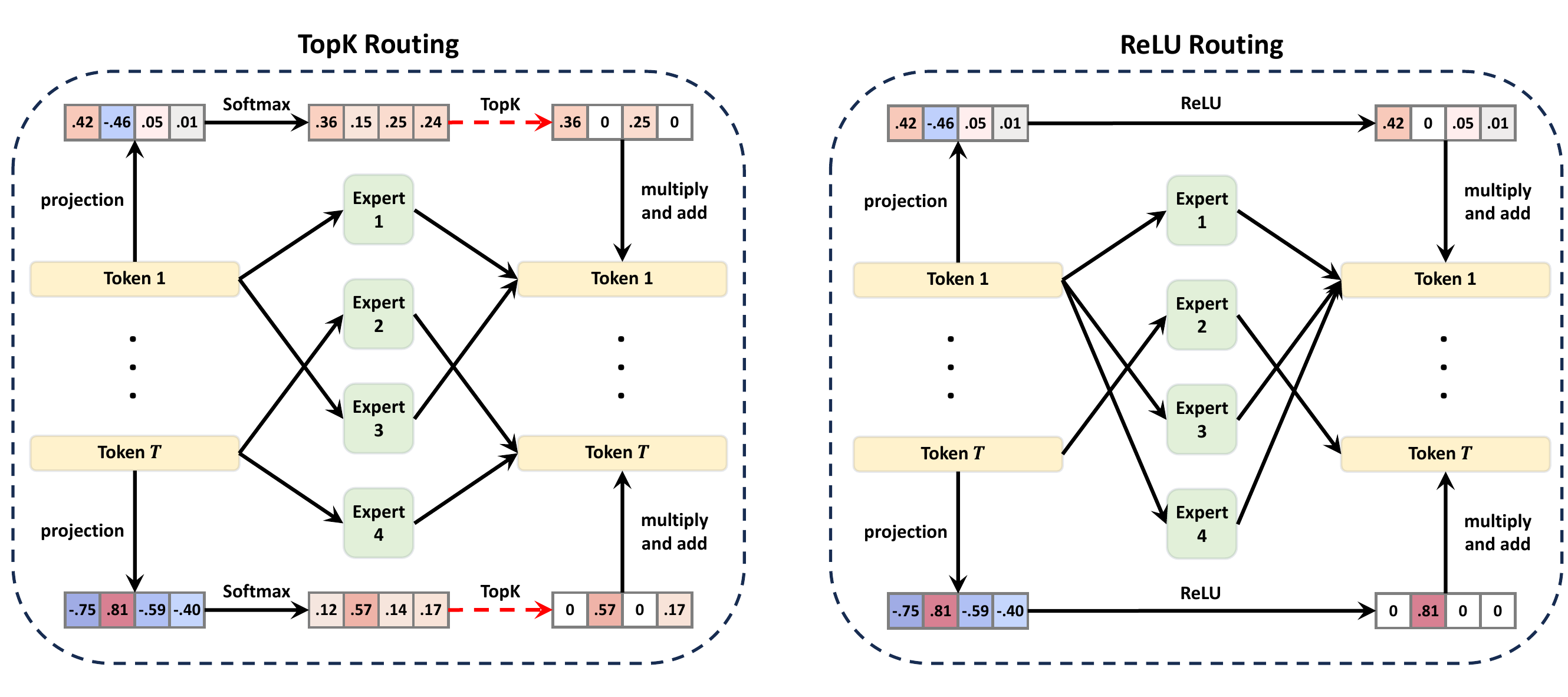}
    \caption{Compute flows of vanilla MoE with TopK routing and ReMoE with ReLU routing. Positive values are shown in orange, and negative values in blue, with deeper colors representing larger absolute values. Zeros, indicating sparsity and computation savings, are shown in white. The red dash arrows in TopK routing indicate discontinuous operations. Compared with TopK routing MoE, ReMoE uses ReLU to make the compute flow fully differentiable.}
    \label{fig:main_graph}
\end{figure}

\section{Preliminaries}
\subsection{MoE for Decoder-Only Transformer}

\newcommand{\ffn}{\text{FFN}}
\newcommand{\topk}{\text{TopK}}
\newcommand{\sm}{\text{Softmax}}

A typical decoder-only Transformer model consists of $L$ layers, each containing a Self-Attention module and a Feed-Forward  Network (FFN) module. MoE modifies this structure by replacing each FFN module with an MoE module, which comprises a small router and several experts $\ffn_1,\dots,\ffn_E$, where each expert is equivalent to the original FFN and $E$ denotes the number of experts.  Given the input $\vx^l=(\vx^l_t)_{t=1}^T\in\R^{T\times d}$ of the layer $l$, where $T$ is the number of tokens in a batch and $d$ is the hidden size, the output $\vy^l=(\vy^l_t)_{t=1}^T$ is computed as:
\begin{align}
    \vy^l_t=\sum_{e=1}^E R(\vx^l_t)_e \ffn_e(\vx^l_t; d_{ffn})
\end{align}
Here, $R(\cdot)$ represents the routing function, and $d_{ffn}$ is the intermediate size of the FFN, typically set to $d_{ffn} = 4d$.

\subsection{TopK Routing}
TopK routing~\citep{shazeer2017outrageously,lepikhin2020gshard,fedus2022switch} is the most commonly used method for defining the routing function $R(\cdot)$. It introduces sparsity in the MoE computation by forcibly zeroing out smaller elements:
\begin{align}
    R(\vx^l_t)=\topk(\sm(\vx^l_t\mW_{l}), k)
\end{align}
where $\mW_l \in \R^{d \times E}$ is the router’s weight matrix, and $\topk(\cdot, k)$ retains the top $k$ largest values while setting the rest to zero. This mechanism allows for skipping the computation of the $\ffn_e$ functions corresponding to the zeroed-out $R(\vx^l_t)_e$ values in both the forward and backward passes.

%% file: src/sec03-method.tex
\section{Our Method: ReMoE}

\subsection{Motivation: From TopK to ReLU}
\newcommand{\relu}{\text{ReLU}}
\newcommand{\xk}{x_{[k]}}

For a given token $\vx=(x_e)_{e=1}^E$ after Softmax, TopK introduces a jump discontinuity at the $k$-th largest value, denoted as $\xk$, by zeroing out the values smaller than $\xk$. This can be expressed as:
\begin{align}
    \topk(\vx, k)_e=x_e\cdot\1\{x_e\ge t(\vx,k)\},\quad t(\vx,k)=\xk
    \label{eqn:topk}
\end{align}

\begin{wrapfigure}{r}{0.5\textwidth}
    \centering
    \includegraphics[width=\linewidth]{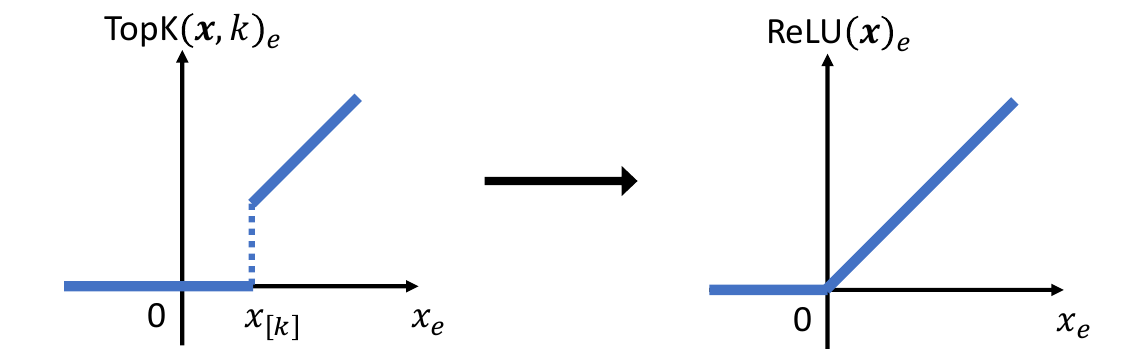}
    \caption{Comparison between TopK and ReLU.}
    \label{fig:topk_relu}
\end{wrapfigure}

where $\1\{\cdot\}$ is the indicator function, returning 1 if the condition is met and 0 otherwise.

As shown in Figure~\ref{fig:topk_relu}, the jump discontinuity can be eliminated by setting the breakpoint $t(\vx,k)\equiv0$, which actually corresponds to the ReLU function:
\begin{align}
    \relu(\vx)_e=x_e\cdot\1\{x_e\ge0\}
\end{align}

At a high level, ReLU improves upon TopK by aligning the breakpoints of all inputs and setting them to 0. This ensures that the output is continuous at 0, where the experts transition between active and inactive. As a result, the training pipeline becomes fully differentiable.

\subsection{Differentiable ReLU Routing}

We define the ReLU routing function as follows:
\begin{align}
    R(\vx^l_t)=\relu(\vx^l_t\mW_l)
    \label{eqn:relu-routing}
\end{align}
with $(1-\frac{k}{E})$ being the desired sparsity of ReLU, where $k$ is the number of active experts and $E$ is the total number of experts. This ensures that the computational cost remains equivalent to that of TopK routing.

In vanilla TopK routers, the Softmax outputs sum to 1, representing the probabilities of selecting each expert, after which TopK eliminates those with lower probabilities. In contrast, ReLU routers discard the Softmax function, relying on ReLU’s naturally non-negative outputs. The outputs of ReLU routers represent the weights assigned to each expert, which can include 0. Instead of hard-coding expert selection with a discontinuous TopK function, ReLU allows the router to learn which experts to activate (i.e., when to produce 0s) in a fully differentiable manner.

Another key difference is that in TopK routing, each token is routed to exactly $k$ experts, whereas in ReLU routing ReMoE, the routing decisions are independent, allowing tokens to be routed to a variable number of experts. This flexibility is advantageous, as not all tokens have the same level of difficulty. ReMoE can allocate more computational resources to more challenging tokens, a dynamic allocation strategy that we explore further in Section~\ref{sec:dynamic_alloc}.

TopK routing introduces a discrete loss function when the set of activated experts changes, whereas ReLU routing remains continuous and fully differentiable. For instance, in a two-expert Top1-routing model, a small weight update that alters the softmax result from \(\vx_1 = (0.51, 0.49)\) to \(\vx_2 = (0.49, 0.51)\) shifts the TopK output from \((0.51, 0)\) to \((0, 0.51)\), creating a discontinuity. In contrast, ReLU routing only changes the activated experts when the routing output is near zero. For example, an output shift from \((0.01, 0)\) to \((0, 0.01)\) remains continuous. \rebuttal{Further details on the stability analysis of these two routers can be found in Appendix~\ref{app:stability}.}

A comparison of the compute flow between ReMoE and MoE is shown in Figure~\ref{fig:main_graph}.

\subsection{\texorpdfstring{Controlling Sparsity via Adaptive $L_1$ Regularization}{Controlling Sparsity via Adaptive L1 Regularization}}

ReMoE controls computational costs by managing the sparsity of the ReLU output, targeting a sparsity level of \((1-\frac{k}{E})\). However, directly training the ReLU router often results in lower sparsity, as the model tends to activate more experts to increase capacity. To meet the desired budget, we need to enforce higher sparsity in the ReLU output.

We achieve this by introducing a regularization loss, $\Ls_{reg}$, to the loss of language model, $\Ls_{lm}$:
\begin{align}
    \Ls &= \Ls_{lm}+\lambda_i \Ls_{reg},
\end{align}
where $\lambda_i$ is an adaptive coefficient based on the current training step $i$.
Initially, we set $\lambda_0$ to a small value and employ a simple zeroth-order algorithm to update it:
\begin{align}
    \lambda_{i+1}&=\lambda_{i}\cdot\alpha^{\text{sign}((1-\frac{k}{E})-S_i)}
    \label{eqn:lambda_update}
\end{align}
Here, $\alpha > 1$ is a preset update multiplier, and $S_i$ denotes the average sparsity of all router outputs at the step $i$:
\begin{align}
    S_i = 1-\frac1{LTE}\sum_{l=1}^L\sum_{t=1}^{T}\sum_{e=1}^{E}\1\{R(\vx^l_t)_e>0\}
\end{align}

The idea behind Equation~\ref{eqn:lambda_update} is that when the average sparsity $S_i$ falls below the target sparsity \((1-\frac{k}{E})\), we increase \(\lambda_i\) by a factor of \(\alpha\), strengthening the regularization and encouraging higher sparsity. Conversely, if the sparsity exceeds the target, \(\lambda_i\) is reduced. We heuristically set $\lambda_0 = 1e^{-8}$ and $\alpha = 1.2$ in all our experiments, and demonstrate the robustness of these hyperparameters in Appendix~\ref{sec:l1_abl}.

The regularization term $\Ls_{reg}$ uses the $L_1$-norm, following prior work \citep{li2022lazy,song2024prosparse}, to effectively encourage sparsity:
\begin{align}
    \Ls_{reg}=\frac1{L T}\sum_{l=1}^{L}\sum_{t=1}^T\norm{R(\vx^l_t)}_1=\frac{1}{LT}\sum_{l=1}^{L}\sum_{t=1}^{T}\sum_{e=1}^{E}R(\vx^l_t)_e
    \label{eqn:l1_reg}
\end{align}
The second equation holds because the output of the ReLU function is non-negative. 

\begin{wrapfigure}{r}{0.5\textwidth}
    \centering

    \includegraphics[width=\linewidth]{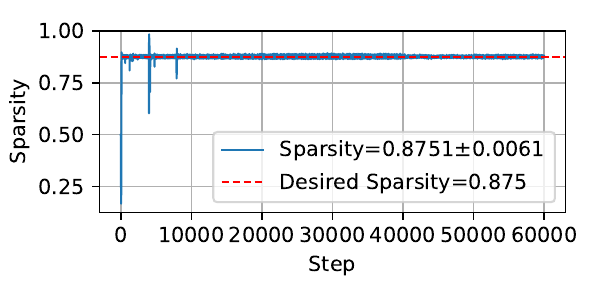}
    \vspace{-0.5cm}
    \caption{The sparsity of ReMoE with $E=8, k=1$ is effectively maintained around the desired target. Sparsity values for all steps are plotted without averaging or sampling. The mean and standard deviation are calculated excluding the first 100 warm-up steps.}
    \label{fig:plot_sparsity}
\end{wrapfigure}

The term $\Ls_{reg}$ represents the average value of all router outputs, including zeros. By taking the derivative of $\lambda_i \Ls_{reg}$, we observe that \rebuttal{the regularization effect adds $\frac{\lambda_i}{L T}$ to the gradient of each non-zero router output, effectively driving the outputs toward zero and enhancing sparsity.}

With this $L_1$ regularization, we can control the sparsity around the desired level of \((1 - \frac{k}{E})\) with only minor fluctuations, as shown in Figure~\ref{fig:plot_sparsity}. Consequently, ReMoE ensures that, on average, tokens are routed to $k$ experts across different layers and tokens, maintaining the same FLOPs as vanilla TopK-routed MoE from a statistical perspective. Our benchmarking results in Appendix~\ref{app:speed} demonstrate that ReMoE can achieve nearly identical training and inference throughputs as conventional MoE, providing an efficient alternative without compromising speed.

\subsection{\texorpdfstring{Integrate Load Balancing into $L_1$ Regularization}{Load Balancing in L1 Regularization}
}

Load imbalance is a significant issue in MoE design, potentially leading to routing collapse~\citep{shazeer2017outrageously,muennighoff2024olmoe} and uneven computational distribution across multiple devices. The $L_1$ regularization in Equation~\ref{eqn:l1_reg} treats the router output for each expert $e$ and each layer $l$ equally, which can contribute to load balancing problems.

\begin{figure}[t]
    \centering
    \begin{subfigure}{0.32\textwidth}
    \centering
    \includegraphics[width=\linewidth]{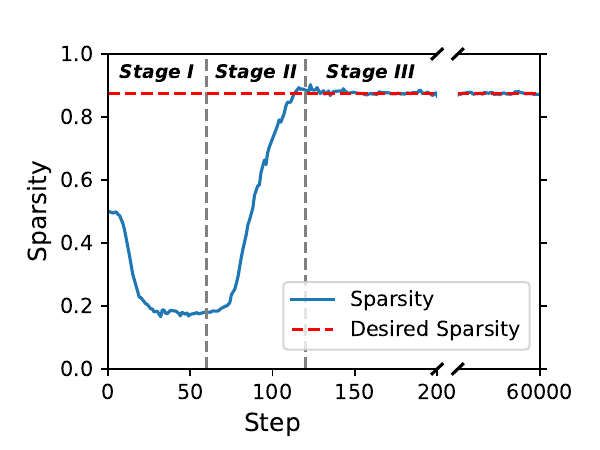}
    \caption{Sparsity $S_i$\\~} 
    \label{fig:three_stage_sparsity}
    \end{subfigure}
    \hfill
    \begin{subfigure}{0.33\textwidth}
    \centering
    \includegraphics[width=\linewidth]{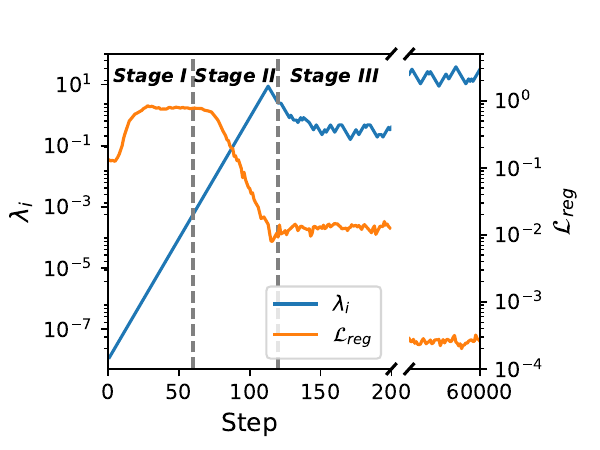}
    \caption{Coefficient term $\lambda_i$ and regularization term $\Ls_{reg}$}
    \end{subfigure}
    \hfill
    \begin{subfigure}{0.33\textwidth}
    \centering
    \includegraphics[width=\linewidth]{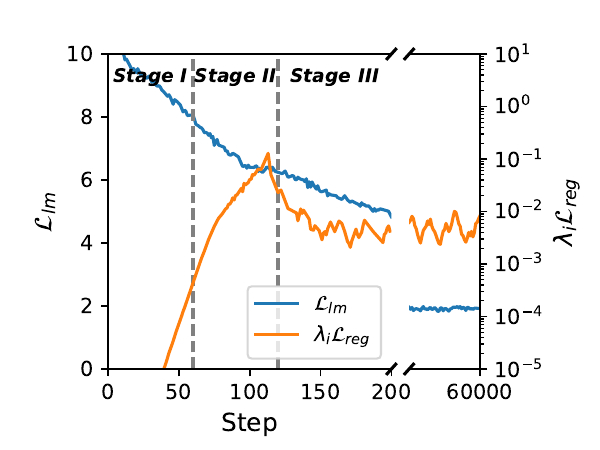}
    \caption{Language model loss $\Ls_{lm}$ and overall regularization $\lambda_i\Ls_{reg}$}    
    \end{subfigure}
    \vspace{-0.2cm}
    \caption{Natural Three Stage Training in ReMoE.}
    \label{fig:three_stage}
    \vspace{-0.4cm}
\end{figure}

To address this, we introduce a load-balancing refinement to the $L_1$ regularization:
\begin{align}
    \Ls_{reg,lb}&=\frac1{L T}\sum_{l=1}^L\sum_{t=1}^{T}\sum_{e=1}^{E}f_{l,e}R(\vx^l_t)_e
    \label{eqn:l1_reg_lb} \\
    f_{l,e}&=\frac{E}{kT}\sum_{t=1}^{T}\1\{R(\vx^l_t)_e>0\}
\end{align}

\vspace{-0.2cm}
Here, \( f_{l,e} \) is non-differentiable and represents the average activation ratio of expert $e$ in layer $l$, relative to the desired ratio \(\frac{k}{E}\). \rebuttal{This serves as a weight for the corresponding router output, modifying the added gradient of non-zero router outputs to $\frac{f_{l,e}\lambda_i}{LT}$. This mechanism penalizes experts receiving more tokens by driving their router outputs toward zero more rapidly.}

Although derived from regularization, this formulation is \textit{identical} to the load-balancing loss in vanilla TopK routing~\citep{fedus2022switch}. In TopK routing, the outputs of Softmax sum to 1, giving the loss a lower bound of 1. In contrast, ReLU routing outputs can be arbitrarily small, making \(\Ls_{reg,lb}\) trivially bounded at 0. Therefore, unlike in MoE, we cannot fix the coefficient \(\lambda_i\) in ReMoE, as this would lead to routing collapse toward 0. Thanks to the adaptive update of \(\lambda_i\), we can balance sparsity control and load balancing within a single formulation, as given in Equation~\ref{eqn:l1_reg_lb}.

Further discussion on load balancing in ReMoE can be found in Section~\ref{sec:w_or_wo_lb}, and we adopt this load-balancing refined $L_1$ regularization in our later experiments.

\subsection{Natural Three-Stage Training in ReMoE}
\label{sec:three_stage}

With the regularization scheme described above, we observe a clear and naturally occurring three-stage separation during the training of ReMoE as is depicted in Figure~\ref{fig:three_stage}.

The first stage is the warm-up stage, or the dense stage. During this stage, $\lambda_i$ is small, while $\Ls_{lm}$ is large and decreases rapidly. Training ReMoE at this stage is nearly equivalent to training its dense counterpart with the same total number of parameters. Each expert processes more than half of the tokens, allowing the experts to diversify from their random initializations.

The second stage is the sparsifying stage, or the dense to sparse stage. At this point, the sparse regularization term $\lambda_i\Ls_{reg}$ becomes significant, causing the ReLU routers to activate fewer experts. This forces the experts to become more diverse without causing an increase in $\Ls_{lm}$.

The third stage is the stable stage, or the sparse stage. In this phase, the sparsity $S_i$ stabilizes at the preset target. During this stage, $\Ls_{lm}$ is optimized while being softly guided along the sparse subspace by $\Ls_{reg}$. Both $\Ls_{reg}$ and $\lambda_i$ change very slowly, with $\Ls_{reg}$ gradually decreasing and $\lambda_i$ gradually increasing. However, the overall regularization term, $\lambda_i\Ls_{reg}$, remains relatively constant. 

\rebuttal{It should be noted that Stages I and II introduce additional computational cost and memory consumption since more experts are activated. However, the time overhead is negligible since they generally require only $\sim$100 iterations ($\sim$0.17\% of the total steps in our setting, benchmarking results are detailed in Appendix~\ref{app:speed}). The memory overhead can be minimized by temporarily reducing the micro-batch size or by employing the activation checkpointing technique that avoids storing intermediate results of activated experts by recomputing them on-the-fly during the backward pass.}

%% file: src/sec04-exp.tex
\section{Experiments}

\subsection{Setup}
\paragraph{Infrastructure}
We leverage Megatron-LM~\citep{shoeybi2019megatron} as our code base and implement ReLU routing as a drop-in replacement for the original TopK routing, supporting all forms of model parallelism:  Data, Tensor, Pipeline, and Expert Parallelism~\citep{shoeybi2019megatron,narayanan2021efficient,korthikanti2023reducing}.

\paragraph{Model Architecture.}
We experiment with the mainstream LLaMA~\citep{touvron2023llama} architecture, featuring grouped query attention (GQA)~\citep{ainslie2023gqa}, SwiGLU ~\citep{shazeer2020glu} activation function, RoPE~\citep{su2024roformer} position embedding, and RMSNorm~\citep{zhang2019root}. The context length is set to 1024, and the batch size is 512. We experiment with three different dense backbone sizes as shown in Table~\ref{tab:model_config}. For vanilla MoE we adopt a load balancing loss of weight $0.01$ following~\cite{fedus2022switch}. For ReMoE we use the adaptive load balancing $L_1$ regularization in Equation~\ref{eqn:l1_reg_lb}.

\paragraph{Training Settings.}
We train the models on The Pile~\citep{gao2020pile}, an 800 GB diverse corpus. All models are trained for 60k steps ($\sim30$B tokens), which exceeds the compute-optimal dataset size predicted by ~\cite{krajewski2024scaling} and is enough to converge. The byte pair encoding (BPE) tokenizer~\citep{sennrich2015neural} is used. We adopt AdamW~\citep{loshchilov2017decoupled} as the optimizer with $\beta_1=0.9,\beta_2=0.999$ with ZeRO optimization~\citep{rajbhandari2020zero}. The learning rate is set to be $5e^{-4}$ with a cosine scheduler. All models are trained with 8 NVIDIA A100 GPUs.

\begin{table}[h]
    \centering
    \begin{threeparttable}
    \resizebox{\linewidth}{!}{
    \begin{tabular}{c|ccccccc}
    \toprule
        \textbf{Size} & \textbf{\#Parameters} & \textbf{hidden\_size} & \textbf{num\_layers} & \textbf{num\_heads} & \textbf{num\_groups} & \textbf{GFLOPs} \\
    \midrule
        Small & 182M & 768 & 12 & 12 & 4 & 995 \\
        Medium & 469M & 1024 & 24 & 16 & 4 & 2873 \\
        Large & 978M & 1536 & 24 & 16 & 4 & 5991 \\
    \bottomrule
    \end{tabular}}
    \end{threeparttable}
    \caption{Configurations for the dense backbones. FLOPs are calculated with a single sequence according to~\cite{narayanan2021efficient}.}
    \label{tab:model_config}
\end{table}

\begin{figure}[t]
    \centering
    
    \begin{minipage}{0.33\textwidth}
        \includegraphics[width=\linewidth]{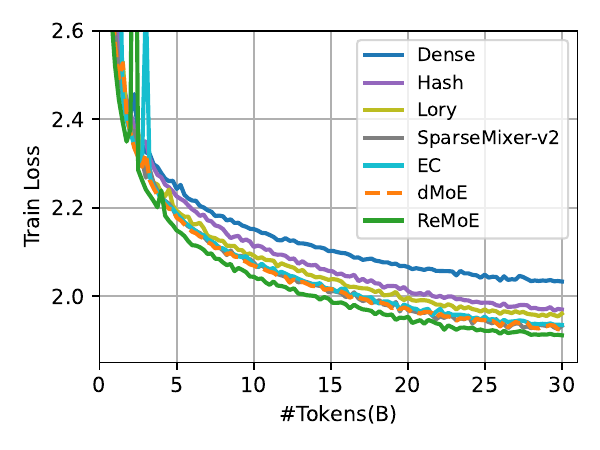}
        \vspace{-0.7cm}
        \caption{\rebuttal{Training curves of different routing methods.}}
        \label{fig:plot_baselines_train}
    \end{minipage}
    \hspace{0\textwidth}
    \begin{minipage}{0.655\textwidth}
    \centering
    \renewcommand{\arraystretch}{1.4}
    \vspace{0.23cm}
    \resizebox{\textwidth}{!}{
    
    \begin{tabular}{c|ccccccc|c}
        \toprule
        \textbf{Model} & \textbf{ARC-c} & \textbf{ARC-e} & \textbf{BoolQ} & \textbf{HellaSwag} & \textbf{LAMBADA} & \textbf{PIQA} & \textbf{RACE} & \textbf{Avg.} \\
        \midrule
        Dense & 19.45 & 43.35 & 54.40 & 28.61 & 31.09 & 61.97 & 28.52 & 38.20 \\
        Hash  & 19.28 & 45.45 & 54.95 & 29.68 & 31.44 & 63.06 & 27.66 & 38.79 \\
        Lory  & \textbf{20.31} & 42.97 & 49.54 & 28.75 & 32.35 & 62.24 & 27.75 & 37.70 \\
        SparseMixer-v2 & 19.80 & \textbf{46.72} & 45.96 & 30.24 & 34.12 & 62.89 & 29.00 & 38.39 \\
        EC    & 18.86 & 42.97 & \textbf{60.21} & 29.14 & 29.26 & 61.92 & 27.37 & 38.53 \\
        dMoE  & 20.05 & 45.16 & 57.83 & 29.83 & 32.97 & \textbf{63.55} & 28.33 & 39.67 \\
        ReMoE & 20.22 & 46.68 & 54.16 & \textbf{30.26} & \textbf{35.94} & \textbf{63.55} & \textbf{29.38} & \textbf{40.03} \\
        \bottomrule
    \end{tabular}
    }
    \vspace{0.10cm}
    \captionof{table}{\rebuttal{Zero-shot accuracy of different routing methods on downstream tasks.}}
    \label{tab:acc_baselines}
    \end{minipage}

\end{figure}

\subsection{Comparison with Other Routing Methods}

We compare ReMoE against the following methods: (i) Token-choice dropless TopK routing (dMoE)~\citep{gale2023megablocks} (ii) Expert-choice TopK routing (EC)~\citep{zhou2022mixture} (iii) Deterministic hash routing (Hash)~\citep{roller2021hash} (iv) Fully-differentiable expert-merging routing (Lory)~\citep{zhong2024lory}\rebuttal{ (v) TopK routing with improved gradient estimate (SparseMixer-v2)}~\citep{liu2024grin}. 

The performance of these methods is evaluated with active parameters $N=182$M and the expert count $E=8$. We fix the active expert count to $k=1$ for straightforward comparison with the dense counterpart. For the Hash method, we use$\mod E$ hashing function. And for Lory, the segment length is set to $256$, following the original paper.

These models are trained on 30B tokens, with the training curves shown in Figure~\ref{fig:plot_baselines_train}, 
We evaluate the zero-shot performance of the trained models on the following downstream tasks: ARC~\citep{clark2018think}; BoolQ~\citep{clark2019boolq}; HellaSwag~\citep{zellers2019hellaswag}; LAMBADA~\citep{paperno2016lambada}; PIQA~\citep{bisk2020piqa}; RACE~\citep{lai2017race}.

The downstream accuracy results are summarized in Table~\ref{tab:acc_baselines}.

Our results show that all MoE models outperform the dense model. Deterministic hash routing performs worse than the learned routing methods. Among the Top-K approaches, token-choice dMoE outperforms expert-choice MoE \rebuttal{and SparseMixer-v2} in evaluation. The differentiable routing method Lory surpasses Hash routing in training but underperforms in downstream tasks, with both methods falling short of the standard Top-K routing. Notably, ReMoE outperforms all methods, including the mainstream Top-K routing, while benefiting from differentiability.

\begin{figure}
    \centering
    \begin{subfigure}{0.32\textwidth}
    \centering
    \includegraphics[width=\linewidth]{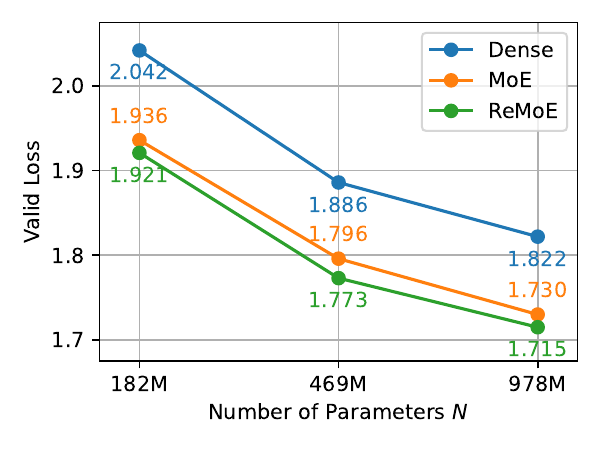}
    \caption{Scaling in $N$}
    \label{fig:plot_n}
    \end{subfigure}
    \hfill
    \begin{subfigure}{0.33\textwidth}
    \centering
    \includegraphics[width=\linewidth]{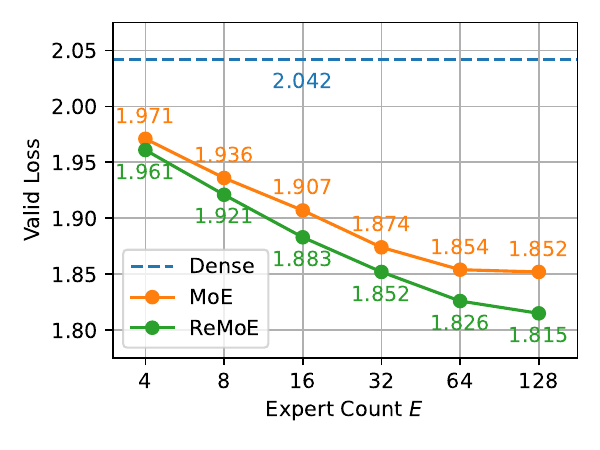}
    \caption{Scaling in $E$}
    \label{fig:plot_e}
    \end{subfigure}
    \hfill
    \begin{subfigure}{0.33\textwidth}
    \centering
    \includegraphics[width=\linewidth]{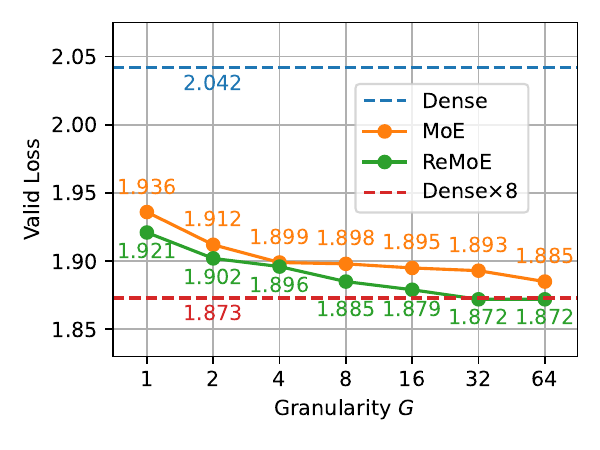}
    \caption{Scaling in $G$}
    \label{fig:plot_g}
    \end{subfigure}
    \caption{Scalability of ReMoE with respect to the number of active parameters ($N$), expert count ($E$), and granularity ($G$). Default config is $N=182\text{M},E=8,G=1,k=1$. The Y-axis represents the validation loss of each model after training on 30B tokens. ReMoE consistently outperforms MoE across all configurations.}
    \label{fig:plot_scalability}
\end{figure}

\subsection{Scalability of ReMoE}
\label{sec:scalability}

In this section, we compare ReMoE with state-of-the-art dMoE (hereinafter referred to simply as MoE) across varying model parameters $N$, expert counts $E$, and granularity levels $G$ to demonstrate its scalability and universal superiority.
Since ReMoE demands more computation in both Stage I and Stage II, we increase the number of training steps for the MoE baseline to match the total computation in each setting, ensuring a more equitable comparison.
We present the final validation losses in Figure~\ref{fig:plot_scalability}, with comprehensive downstream evaluation results available in Appendix~\ref{sec:full_evaluation}.

\paragraph{\rebuttal{Scaling in active parameters $N$.}}
To assess scalability with respect to the number of parameters $N$, we fix $E=8$ and $k=1$, \rebuttal{while varying active parameters $N$ from 182M to 975M, corresponding to the dense counterpart configurations in Table~\ref{tab:model_config}. The total parameters are 777M, 2.58B, 5.73B respectively}. The results, shown in Figure~\ref{fig:plot_n}, indicate that ReMoE consistently outperforms MoE across all model sizes. The performance gap does not diminish as the model size increases, suggesting that ReMoE maintains its advantage at larger scales.

\paragraph{Scaling in expert count $E$.}
In this experiment, we fix the number of parameters at $N=182$M and set the number of active experts $k=1$, while varying the total number of experts $E$ from 4 to 128. The scaling curve in Figure~\ref{fig:plot_e} reveals that ReMoE consistently outperforms the standard MoE across all configurations of $E$.

Moreover, a key observation is the steeper slope in ReMoE’s performance as $E$ increases, compared to MoE. This suggests that ReMoE scales more effectively with the number of experts and derives greater benefits from larger expert pools. ReMoE’s differentiable routing strategy appears better suited for leveraging large expert groups, leading to significant improvements in model expressivity and generalization.

\paragraph{Scaling in granularity $G$.}
We also evaluate ReMoE and MoE in fine-grained settings. Fine-grained MoE~\citep{dai2024deepseekmoe,krajewski2024scaling} with granularity $G$ is constructed by dividing each expert into $G$ smaller experts, as formulated below:
\begin{align}
    \vy^l_t&=\sum_{e=1}^{EG} R(\vx^l_t)_e \ffn_e(\vx^l_t; d_{ffn}/G) \\ R(\vx^l_t)&=\topk(\sm(\vx^l_t\mW_{l}), kG)
\end{align}
Fine-grained MoE outperforms vanilla MoE from a scaling law perspective~\citep{krajewski2024scaling} and has been adopted in subsequent works~\citep{dai2024deepseekmoe,tan2024scattered,muennighoff2024olmoe}. For fine-grained ReMoE, the routing function remains identical to Equation~\ref{eqn:relu-routing}, and the target sparsity is still $(1-\frac{k}{E})$. The only distinction lies in the shape of the weight matrix, with $\mW_l\in\R^{d\times EG}$.

We conduct experiments with $N=182$M and $E=8$, varying $G$ from 1 to 64 for both fine-grained MoE and fine-grained ReMoE. In addition to comparing these models against the dense baseline with the same number of active parameters, we also evaluate their dense counterpart with the same total number of parameters. This is achieved by expanding the intermediate size of the FFN by a factor of $E$, which we denote as \textit{Dense$\times$8}. This configuration represents the strict upper bound for MoE and ReMoE, as it is equivalent to a Mixture-of-Experts with all experts activated~\citep{dai2024deepseekmoe}.

As illustrated in Figure~\ref{fig:plot_g}, fine-grained ReMoE consistently outperforms fine-grained MoE. Moreover, fine-grained ReMoE of $G=32$ and $G=64$ reach the performance of the theoretical upper bound, \textit{Dense$\times$8}, while requiring significantly fewer FLOPs during both training and inference. In contrast, fine-grained MoE is unable to match in all settings, making ReMoE a more efficient and effective choice.

%% file: src/sec05-discussion.tex
\section{Discussion}

\subsection{Dynamic Expert Allocation in ReMoE}
\label{sec:dynamic_alloc}

\begin{wrapfigure}{r}{0.4\linewidth}
    \centering
    \includegraphics[width=\linewidth]{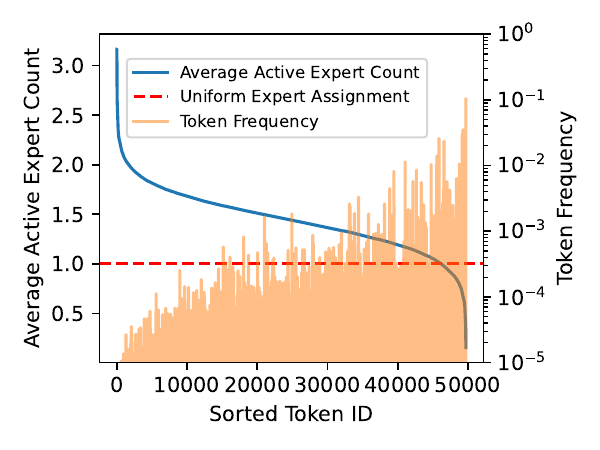}
    \caption{Correlation between expert allocation and token frequency in ReMoE. X-axis is sorted by average active expert count and token frequency is in log-scale.}
    \label{fig:dynamic_token}
\end{wrapfigure}

In ReMoE, each token dynamically activates a subset of experts, allowing the model to adaptively allocate resources. We evaluate the performance of the $N=182\text{M}, E=8, k=1$ ReMoE model and analyze the relationship between token frequency and the average number of active experts. As illustrated in Figure~\ref{fig:dynamic_token}, the model tends to assign a higher number of experts to rarer tokens, such as \texttt{'©'}, \texttt{'OTAL'}, and \texttt{'@\#'}, while reducing the number of active experts for more frequent tokens like \texttt{' '}, \texttt{'\textbackslash n'}, and \texttt{'the'}.

This adaptive behavior mirrors the principles of a Huffman tree~\cite{huffman1952method}, where more frequent symbols are assigned shorter codes, and rarer symbols are assigned longer codes. Similarly, ReMoE tends to ``cluster on'' common tokens by activating fewer experts, effectively compressing the ``representation'' of these frequent tokens. In contrast, for rarer tokens, ReMoE activates a more diverse set of experts, ``encoding'' them as a richer linear combination at the expert level. This suggests that ReMoE learns to dynamically allocate computational resources, achieving an efficient balance between resource usage and the model's capacity, optimizing performance under a constrained expert budget. \rebuttal{Dynamic expert allocation is also evident at the domain level, as detailed in Appendix~\ref{app:dynamic_domain}.}

\subsection{The Role of Load Balancing in ReMoE}
\label{sec:w_or_wo_lb}

\begin{figure}[t]
    \centering
    \begin{subfigure}{0.28\textwidth}
        \centering
        \includegraphics[width=\linewidth]{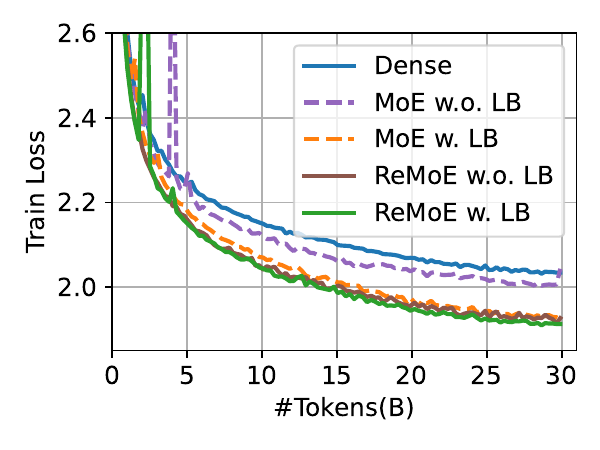}
        \caption{Training curves of MoE and ReMoE with and without load balancing}
        \label{fig:plot_lb}
    \end{subfigure}
    \hfill
    \begin{subfigure}{0.2\textwidth}
        \centering
        \includegraphics[width=\linewidth]{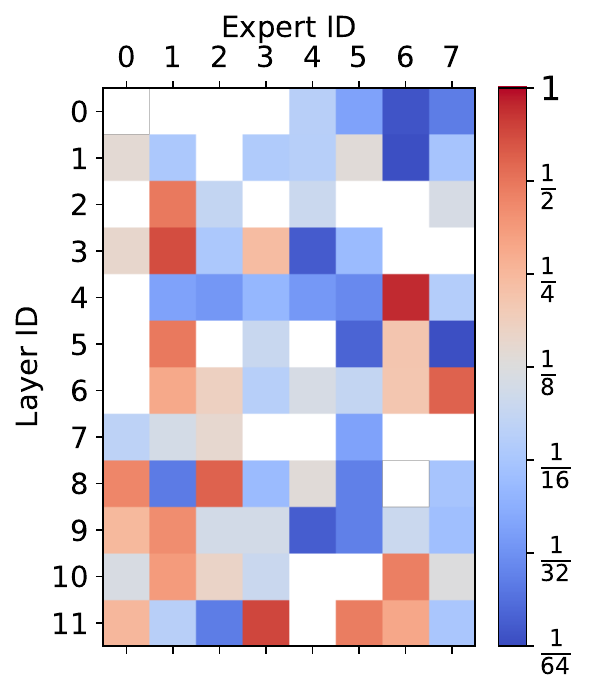}
        \caption{Average routed tokens ratio of ReMoE w.o. LB}
        \label{fig:dynamic_layers_nolb}
    \end{subfigure}
    \hfill
    \begin{subfigure}{0.2\textwidth}
        \centering
        \includegraphics[width=\linewidth]{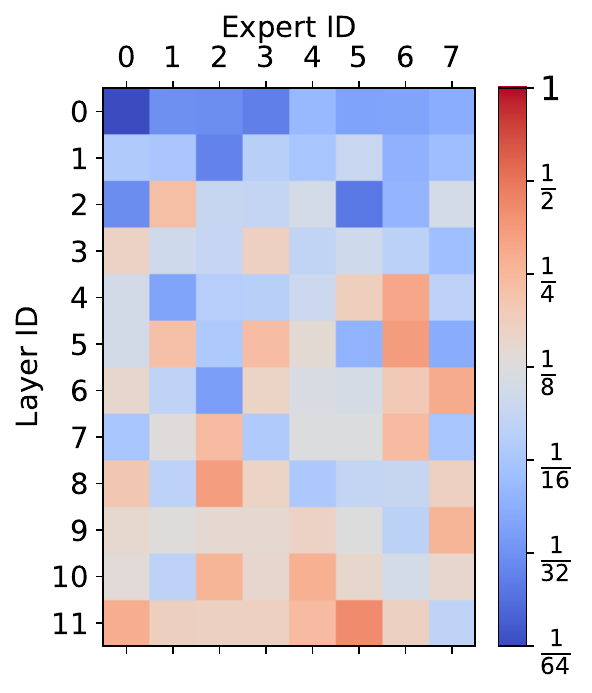}
        \caption{Average routed tokens ratio of ReMoE w. LB}
        \label{fig:dynamic_layers_lb}
    \end{subfigure}
    \hfill
    \begin{subfigure}{0.28\textwidth}
        \centering
        \includegraphics[width=\linewidth]{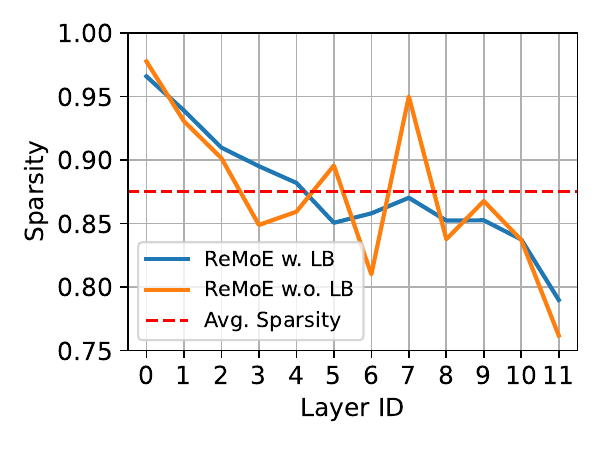}
        \caption{Sparsity across different layers in ReMoE\\~}
        \label{fig:sparsity_per_layer}
    \end{subfigure}
    \caption{Observations on the role of load balancing in MoE and ReMoE. White squares in (b) represent inactive experts with fewer than 1/64 tokens routed to them.}
\end{figure}

Load imbalance can lead to routing collapse in the vanilla TopK-routed MoE, where the router tends to assign the same expert to all inputs, in which scenario the training objective becomes continuous and fully differentiable. As is shown in Figure~\ref{fig:plot_lb}, there is a significant performance gap between MoE models with and without load balancing (LB).

While in ReLU routing, thanks to its differentiablity, even applying the \(L_1\) regularization from Equation~\ref{eqn:l1_reg} without load balancing yields comparable results with a well-tuned MoE with LB. However, some experts in ReMoE without LB remain inactive, illustrated as white squares in Figure~\ref{fig:dynamic_layers_nolb} which shows the heat map of the \textit{average routed tokens ratio} (i.e., the fraction of tokens routed to the $e$-th expert in the $l$-th layer) over 50M tokens in test set. This inactivity can limit the model's capacity. 

When load balancing is incorporated into the refined \(L_1\) regularization (Equation~\ref{eqn:l1_reg_lb}), the experiments show a more even distribution of token assignments across experts, with all experts being utilized, as shown in Figure~\ref{fig:dynamic_layers_lb}. The final loss in ReMoE decreases after introducing load balancing.

Besides, we observe ReMoE with LB can produce a smoother sparsity distribution across layers as depicted in Figure~\ref{fig:sparsity_per_layer}. This is because \(f_{l,e}\) is computed based on the absolute number of routed tokens, meaning denser layers receive stronger penalties.

Note that even ReMoE with load balancing (LB) does not yield a perfectly even distribution. However, the trade-off between load balancing and performance can be easily adjusted by modifying the \(L_1\) regularization in Equation~\ref{eqn:l1_reg_lb}. For instance, changing \(f_{l,e}\) to \(f_{l,e}^2\) would make the model more sensitive to load imbalance. Additionally, device-level load balancing techniques, as proposed in ~\cite{dai2024deepseekmoe}, could also be employed. Since load imbalance in ReMoE does not lead to severe routing collapse, it primarily becomes a hardware utilization issue. As such, we leave the exploration of these variants for future work.

\begin{figure}[h]
    \centering
    \begin{subfigure}{\textwidth}
    \centering
    \includegraphics[width=\linewidth]{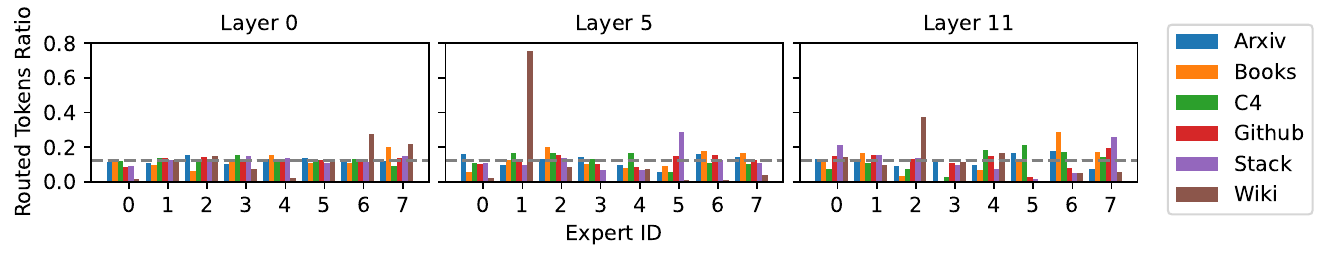}
    \caption{Domain specialization of MoE}
    \label{fig:expert_utilize_moe}
        
    \end{subfigure}
    \begin{subfigure}{\textwidth}
    \centering
    \includegraphics[width=\linewidth]{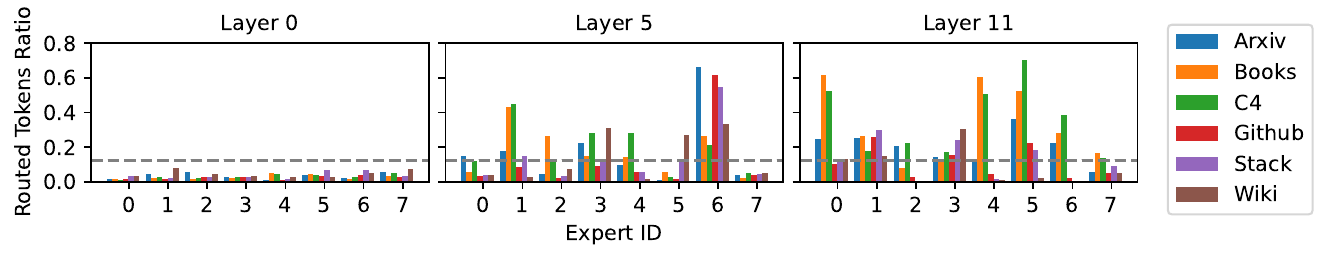}
    \caption{Domain specialization of ReMoE}
    \label{fig:expert_utilize}
    \end{subfigure}
    \caption{Average routed tokens ratio for MoE and ReMoE across 12 layers and 8 experts in different domains. The gray dashed lines indicate uniform distribution. ReMoE shows stronger domain specialization.}
    \label{fig:specilization}
\end{figure}

\subsection{Domain Specialization in ReMoE}

The differentiability and dynamic allocation strategy of ReMoE facilitates the development of diverse experts that specialize in different domains. This allows the router to effectively perform ensemble learning by leveraging the expertise of various experts, as demonstrated in our experiments.

In Figure~\ref{fig:specilization}, we plot the average routed tokens ratio across different experts, layers, and domains—namely Arxiv, Books, C4, Github, Stackexchange, and Wikipedia—for MoE and ReMoE models with $N=182\text{M}, E=8$. We focus on the first, middle, and last layers (with IDs 0, 5, and 11). The results for most experts in MoE (Figure~\ref{fig:expert_utilize_moe}) show a roughly uniform distribution across all domains. In contrast, experts in ReMoE (Figure~\ref{fig:expert_utilize}) exhibit clear domain specialization, being activated with varying frequencies across different domains. For example, more than half of the tokens from Arxiv, Github, and StackExchange—domains that emphasize structured, non-natural languages like LaTeX and Python—are routed to Expert 6 in Layer 5, significantly more than in other domains. A more detailed result of domain specialization can be found in Appendix~\ref{app:domain}.

%% file: src/sec06-related.tex
\section{Related Works}

\subsection{Mixture-of-Experts}

Mixture-of-Experts (MoE) was initially proposed in the early 1990s~\citep{jacobs1991adaptive,jordan1994hierarchical} and later introduced into large-scale neural networks as a sparse submodule for efficiency~\citep{shazeer2017outrageously}. Advances like GShard~\citep{lepikhin2020gshard} and Switch Transformer~\citep{fedus2022switch} integrated sparse MoE into Transformer models, achieving significant results. More recently, MoE has been used in commercial-scale language models such as Mixtral-8x7B~\citep{jiang2024mixtral}, DeepSeekMoE 16B~\citep{dai2024deepseekmoe}, and Snowflake Arctic 17B~\citep{snowflake2024arctic}.

\subsection{Routing Mechanisms in MoE}

Various routing methods have been developed for expert selection. Static routers, such as BASE~\citep{lewis2021base}, use predefined rules like combinatorial optimization, while Hash routing~\citep{roller2021hash} relies on deterministic hash functions, and THOR~\citep{zuo2021taming} assigns experts randomly with regularization. Learned routers adaptively select experts based on token input, using approaches like REINFORCE~\citep{bengio2013estimating,schulman2015gradient,clark2022unified} for reinforcement learning, and TopK routing~\citep{shazeer2017outrageously,zhou2022mixture} for token or expert selection, though TopK introduces discontinuities that hinder gradient estimation.
\subsection{Differentiable Mixture-of-Experts}

Recent work on fully differentiable MoE models addresses the challenges of discrete optimization, basically through token merging and expert merging approaches. Soft MoE~\citep{puigcerver2023sparse} uses token merging, assigning fixed slots to each expert as a linear combination of input tokens. SMEAR~\citep{muqeeth2023soft} merges experts into an ensemble via weighted averaging. However, both methods require a full probability map of input tokens, making them unsuitable for autoregressive models. Lory~\citep{zhong2024lory} preserves autoregressiveness by segmenting sentences to merge experts but underperforms compared to TopK routing.

%% file: src/sec07-conclusion.tex
\section{Conclusion}
In this paper, we propose ReMoE, a fully differentiable MoE architecture with ReLU routing. \rebuttal{The simple yet effective ReLU routing function acts as a drop-in replacement for the conventional TopK+Softmax routing, offering (i) continuity and differentiability, and (ii) dynamic expert allocation across tokens and layers.} With the adaptive load balancing $L_1$ regularization, ReMoE universally outperforms TopK-routed MoE across various model sizes, expert counts, and levels of granularity, demonstrating sharper performance gains as the number of experts scales.

\newpage
\section*{Acknowledgment}

The authors gratefully acknowledge Chao Du and Tianyu Pang for the insightful discussions. This work was supported by the NSFC Project (No.~62376131), Tsinghua Institute for Guo Qiang, and the High Performance Computing Center, Tsinghua University. J.Z is also supported by the XPlorer Prize.

%% file: src/appendix.tex
\section{Stability analysis of TopK and ReLU}
\label{app:stability}

We introduce two metrics, ``flip rate" and ``flip count", to evaluate the routing stability:

\begin{align}
    &\text{flip rate}=\frac{\sum_{l=1}^L \norm{\text{vec}(\mM^l_{i} - \mM^l_{i-1})}_1}{LTE} \\
    &\text{flip count}=E\times\text{flip rate}
\end{align}

where $\mM^l_i\in\R^{T\times E}$ denotes the 0-1 mask matrix of the output of the router at layer $l$ and training step $i$, computed using a \textit{fixed} calibration set of tokens.

The metric ``flip rate" represents the percentage of expert activation states that change (from active to inactive or conversely) in a single update, while ``flip count" indicates the average number of experts whose activation states change.

We measure the two metrics on MoE and ReMoE with $N=$182M and $E\in\{8,16,32\}$ training for 10B tokens. The results are presented in Figure~\ref{fig:flip}, indicating that the ReLU router is more stable than the TopK router:

\begin{figure}[h]
    \centering
    \begin{subfigure}{0.48\textwidth}
    \centering
    \includegraphics[width=\linewidth]{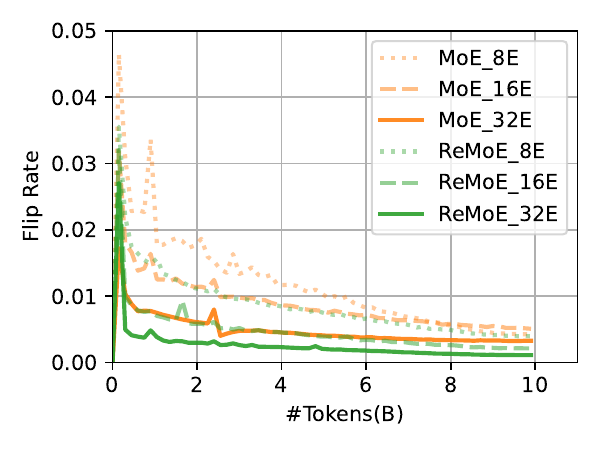}
    \end{subfigure}
    \hfill
    \begin{subfigure}{0.48\textwidth}
    \centering
    \includegraphics[width=\linewidth]{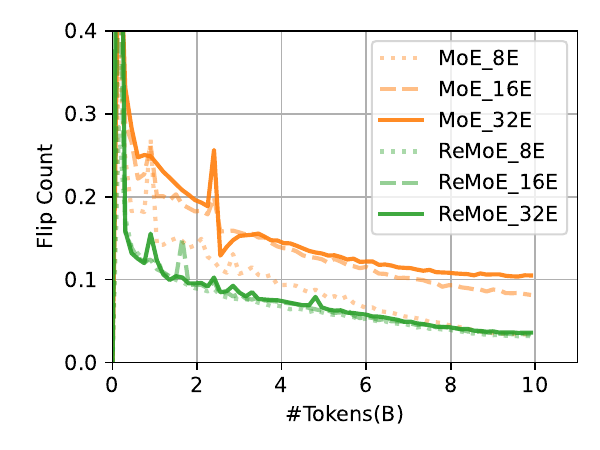}
    \end{subfigure}
    
    \caption{Flip rate and flip count of MoE and ReMoE}
    \label{fig:flip}
\end{figure}

When $E=8$, we find the flip rate of MoE is higher than ReMoE, though the gap narrows as training progresses and the learning rate decreases. While for $E=16$ and $E=32$, the flip rate of MoE remains consistently $2-3\times$ higher compared to ReMoE throughout training. 

Moreover, the flip count of ReMoE is invariant with respect to $E$, whereas the flip count of MoE is highly sensitive to the total number of experts and keeps increasing as $E$ grows.

Notably, the flips in TopK-routed MoE are discontinuous (e.g.$(0.51,0)\rightarrow(0,0.51)$), while those in ReLU-routed ReMoE are continuous(e.g.$(0.01,0)\rightarrow(0,0.01)$), further underscoring the superiority of the ReLU router.

\section{\texorpdfstring{Insensitivity to $\lambda_0$ and $\alpha$}{Insensitivity to lambda\_0 and alpha}}
\label{sec:l1_abl}

\begin{table}[h]
    \centering
    \begin{minipage}{0.48\textwidth}
        \centering
        \resizebox{\textwidth}{!}{%
        \begin{threeparttable}
        \begin{tabular}{cccccc}
            \toprule
             $\lambda_0$ & $1e^{-16}$ & $1e^{-12}$ & $1e^{-8}$ & $1e^{-4}$ & $1$ \\
             \midrule
             Valid Loss & 2.031 & 2.029 & 2.032 & 2.036 & 2.032 \\
             Settling time & 138 & 136 & 110 & 55 & 92$^\dagger$\\
             \bottomrule
        \end{tabular}
        \begin{tablenotes}
            \item[$\dagger$] Overshoot observed in 8-92 steps.
        \end{tablenotes}
        \end{threeparttable}
        }
        \caption{Valid loss and settling time for different values of $\lambda_0$ with $\alpha=1.2$.}
        \label{tab:abl_lambda}
    \end{minipage}%
    \hfill%
    \begin{minipage}{0.48\textwidth}
        \centering
        \resizebox{\textwidth}{!}{%
        \begin{threeparttable}
        \begin{tabular}{cccccc}
            \toprule
             $\alpha$ & 1.05 & 1.1 & 1.2 & 1.3 & 1.5 \\
             \midrule
             Valid Loss & 2.033 & 2.028 & 2.032 & 2.029 & 2.057$^*$ \\
             Settling time & 414 & 211 & 110 & 80 & 52 \\
             \bottomrule
        \end{tabular}
        \begin{tablenotes}
            \item[$*$] A large oscillation amplitude in sparsity is observed.
        \end{tablenotes}
        \end{threeparttable}
        }
        \caption{Valid loss and settling time for different values of $\alpha$ with $\lambda_0=1e^{-8}$.}
        \label{tab:abl_alpha}
    \end{minipage}
\end{table}

The ReMoE adaptation algorithm in Equation~\ref{eqn:lambda_update} includes two hyperparameters: $\lambda_0$ and $\alpha$. \textit{Settling time}, defined as the total number of steps required in Stage I and Stage II (as outlined in Section~\ref{sec:three_stage}), is influenced by these parameters. For all experiments, we set $\lambda_0 = 1e^{-8}$ and $\alpha = 1.2$, but we show that performance remains stable as long as $\lambda_0$ is small and $\alpha$ is close to 1.

Our experiments with $N=182$M, $E=8$, $G=1$, and $k=1$ ReMoE models trained for 20k steps ($\sim$10B tokens) reveal only minor variations in validation loss for different $\lambda_0$ values (Table~\ref{tab:abl_lambda}) and $\alpha$ values (Table~\ref{tab:abl_alpha}), except for $\alpha=1.5$ which caused rapid regularization changes and excessive oscillation. Besides, although different $\lambda_0$ and $\alpha$ values affect settling time, the impact is minor compared to the overall training steps, proving the insensitivity.

\section{Performance for Longer Training}

We conduct experiments of training MoE and ReMoE for a longer duration. We experiment with $N=$469M, $E=8$, $k=1$ and train the models with a batch size of 4M tokens and training over 120B tokens. The results, as shown in Table~\ref{tab:long}, indicate that the superiority of ReMoE persists in longer training.

\begin{table}[h]
    \centering
    \resizebox{\textwidth}{!}{
    \begin{tabular}{c|c|ccccccc|c}
    \toprule
        Model & Valid Loss & ARC-c & ARC-e & BoolQ & HellaSwag & LAMBADA & PIQA & RACE & Avg. \\
        \midrule
        MoE & 1.716 & 23.62 & 52.40 & 53.94 & 35.43 & 43.64 & 68.34 & \textbf{31.48} & 44.12 \\
        ReMoE & \textbf{1.689} & \textbf{25.34} & \textbf{55.22} & \textbf{55.96} & \textbf{36.76} & \textbf{45.82} & \textbf{68.93} & 30.43 & \textbf{45.49} \\\bottomrule
    \end{tabular}}
    \caption{Performance of training $N=$469M, $E=8$, $k=1$ models for 120B tokens.}
    \label{tab:long}
\end{table}

\section{Speed Comparison of ReMoE and MoE}
\label{app:speed}

We measure the end-to-end training time for MoE and ReMoE with models of $N=$469M training over 120B tokens. The time consumption across stages is summarized in Table~\ref{tab:end_to_end}. We find Stage I and Stage II account for $\sim$1.02\% of the total training time and incur $\sim$0.58\% overhead.

\begin{table}[h]
    \centering
    \begin{tabular}{c|ccc|c}
    \toprule
         Model & Stage I & Stage II & Stage III & Total\\
         \midrule
         MoE& 0.12 & 0.41 & 119.12 & 119.65\\
         ReMoE& 0.32 & 0.91 & 119.25 & 120.48 \\
         \bottomrule
    \end{tabular}
    \caption{End-to-end training time comparison across stages (in hours). The time is measured on $N=$ 469M, $E=8$, $k=1$ models training over 120B tokens.}
    \label{tab:end_to_end}
\end{table}

\vspace{0.7cm}
\begin{table}[h]
    \centering
    \resizebox{\textwidth}{!}{
    \begin{tabular}{c|c|c|cc|cc}
        \toprule
        \textbf{\# Parameters} & \textbf{TP} & \textbf{Model} & \textbf{Train TFLOPS} & \textbf{Train Diff.} & \textbf{Infer TFLOPS} & \textbf{Infer Diff.}\\
        \midrule
        \multirow{2}{*}{182M} & \multirow{2}{*}{1} & MoE & 103.49 & \multirow{2}{*}{↑1.82\%} & 78.47 & \multirow{2}{*}{↑2.19\%}\\
                              &                    & ReMoE & 105.38 & & 80.19 & \\
        \midrule
        \multirow{2}{*}{469M} & \multirow{2}{*}{1} & MoE & 138.58 & \multirow{2}{*}{↓1.37\%} & 107.52 & \multirow{2}{*}{↑3.89\%} \\
                              &                    & ReMoE & 136.69 & & 111.71 & \\
        \midrule
        \multirow{2}{*}{978M} & \multirow{2}{*}{1} & MoE & 160.46 & \multirow{2}{*}{↓1.77\%}   & 153.11 & \multirow{2}{*}{↓0.23\%} \\
                              &                    & ReMoE & 157.61 & & 152.76 & \\
        \midrule
        \multirow{2}{*}{978M} & \multirow{2}{*}{2} & MoE & 133.40 & \multirow{2}{*}{↓0.68\%} &  118.55 & \multirow{2}{*}{↓1.08\%} \\
                              &                    & ReMoE & 132.49& & 117.27& \\
        \midrule
        \multirow{2}{*}{978M} & \multirow{2}{*}{4} & MoE & 103.61 & \multirow{2}{*}{↓2.29\%}    &   85.96 & \multirow{2}{*}{↑2.33\%} \\
                              &                    & ReMoE & 101.23 & & 87.96 & \\
        \bottomrule
    \end{tabular}}
    \caption{Throughput comparison between TopK-routed MoE and ReLU-routed ReMoE models. TP indicates the tensor parallel size. Train Diff. and Infer Diff. indicate the relative TFLOPS difference of ReMoE compared to MoE, where ↑ denotes ReMoE is faster, and ↓ denotes it is slower.}
    \label{tab:speed}
\end{table}

We further measure the throughput of ReMoE against TopK-routed MoE across different model sizes and tensor parallel sizes during Stage III. The results, presented in Table~\ref{tab:speed}, indicate that ReMoE achieves comparable training and inference speeds with MoE, with a minor deviation ranging from $-2.29\%$ to $+3.89\%$. This speed consistency is desirable, as ReMoE introduces only a minimal modification to the standard MoE architecture by adjusting the routing function, thereby avoiding additional computational overhead.

\section{Downstream Evaluation Results}
\label{sec:full_evaluation}

This section provides the detailed downstream evaluation results for the main experiments of scalability of ReMoE in Section~\ref{sec:scalability} and ablations on load balancing in Section~\ref{sec:w_or_wo_lb}.

\subsection{\texorpdfstring{Scaling in Active Parameters \( N \)}{Scaling in Active Parameters N}}
The downstream evaluation results for scaling with respect to the parameter count \( N \), as discussed in Section~\ref{sec:scalability}, are presented in Table~\ref{tab:scale_n}. These results highlight the performance comparison with increasing model parameters.

\begin{table}[h]
    \centering
    \resizebox{\textwidth}{!}{
    \begin{tabular}{cc|ccccccc|c}
        \toprule
        \textbf{Model} & $N$ & \textbf{ARC-c} & \textbf{ARC-e} & \textbf{BoolQ} & \textbf{HellaSwag} & \textbf{LAMBADA} & \textbf{PIQA} & \textbf{RACE} & \textbf{Avg.} \\
        \midrule
        \multirow{3}{*}{Dense} 
        & 182M & 19.45 & 43.35 & 54.40 & 28.61 & 31.09 & 61.97 & 28.52 & 38.20 \\
        & 469M & 21.50 & 49.12 & 56.88 & 31.12 & 36.74 & 64.47 & 30.53 & 41.48 \\
        & 978M & 21.93 & 50.88 & \textbf{60.24} & 32.42 & 41.06 & 67.46 & \textbf{31.77} & 43.68 \\ 
        \midrule
        \multirow{3}{*}{MoE} 
        & 182M & 20.82 & 45.03 & 57.55 & 29.84 & 31.81 & 63.28 & 28.42 & 39.53 \\
        & 469M & 23.63 & 52.40 & 53.94 & 32.43 & 43.64 & 68.34 & 31.48 & 43.69 \\
        & 978M & 23.81 & 52.90 & 58.90 & 35.01 & \textbf{44.42} & 67.90 & 31.48 & 44.91 \\ 
        \midrule
        \multirow{3}{*}{ReMoE} 
        & 182M & 20.22 & 46.68 & 54.16 & 30.26 & 35.94 & 63.55 & 29.38 & 40.03 \\
        & 469M & 21.67 & 53.16 & 58.75 & 33.80 & 40.66 & 67.95 & 31.20 & 43.88 \\
        & 978M & \textbf{24.06} & \textbf{55.26} & 57.28 & \textbf{35.93} & \textbf{44.42} & \textbf{68.99} & 30.43 & \textbf{45.20} \\ 
        \bottomrule
    \end{tabular}}
    \caption{Downstream results of scaling in active parameters $N$.}
    \label{tab:scale_n}
\end{table}

\subsection{\texorpdfstring{Scaling in Expert Count \( E \)}{Scaling in Expert Count E}}
Table~\ref{tab:scale_e} contains the downstream evaluation results for scaling with respect to the expert count \( E \), as examined in Section~\ref{sec:scalability}. This analysis illustrates how varying the number of experts influences the overall model effectiveness of MoE and ReMoE.

\begin{table}[h]
    \centering
    \resizebox{\textwidth}{!}{
    \begin{tabular}{cc|ccccccc|c}
        \toprule
        \textbf{Model} & $E$ & \textbf{ARC-c} & \textbf{ARC-e} & \textbf{BoolQ} & \textbf{HellaSwag} & \textbf{LAMBADA} & \textbf{PIQA} & \textbf{RACE} & \textbf{Avg.} \\
        \midrule
        Dense & - & 19.45 & 43.35 & 54.40 & 28.61 & 31.09 & 61.97 & 28.52 & 38.20 \\
        \midrule
        \multirow{6}{*}{MoE} 
        & 4   & 20.73 & 44.49 & 59.63 & 29.14 & 31.40 & 63.33 & 29.19 & 39.70 \\
        & 8   & 20.82 & 45.03 & 57.55 & 29.84 & 31.81 & 63.28 & 28.42 & 39.53 \\
        & 16  & 20.90 & 45.29 & 46.36 & 30.50 & 33.22 & 64.96 & 28.33 & 38.50 \\
        & 32  & 19.54 & 47.35 & 52.29 & 31.12 & 35.63 & 64.25 & 28.23 & 39.77 \\
        & 64  & 19.88 & 46.63 & \textbf{60.06} & 31.47 & 36.33 & 65.07 & 28.04 & 41.06 \\
        & 128 & \textbf{20.99} & 47.69 & 56.73 & 32.00 & 36.62 & 65.67 & 28.04 & 41.10 \\
        \midrule
        \multirow{6}{*}{ReMoE} 
        & 4   & 19.88 & 46.46 & 57.43 & 29.64 & 33.57 & 62.95 & 27.66 & 39.66 \\
        & 8   & 20.22 & 46.68 & 54.16 & 30.26 & 35.94 & 63.55 & 29.38 & 40.03 \\
        & 16  & 20.90 & 49.28 & 53.36 & 30.85 & 37.09 & 65.83 & \textbf{30.05} & 41.05 \\
        & 32  & 20.56 & 48.11 & 59.54 & 31.42 & 37.84 & 65.18 & 28.42 & 41.58 \\
        & 64  & 20.82 & 50.51 & 57.80 & 32.17 & 36.74 & 65.78 & 27.46 & 41.61 \\
        & 128 & 19.97 & \textbf{51.05} & 56.97 & \textbf{32.40} & \textbf{37.92} & \textbf{66.70} & 29.86 & \textbf{42.12} \\
        \bottomrule
    \end{tabular}}
    \caption{Downstream results of scaling in expert count $E$.}
    \label{tab:scale_e}
\end{table}

\subsection{\texorpdfstring{Scaling in Granularity \( G \)}{Scaling in Granularity G}}
The downstream evaluation results for scaling with respect to the granularity \( G \) are shown in Table~\ref{tab:scale_g}, based on the experiments in Section~\ref{sec:scalability}. These results demonstrate the superiority of fine-grained ReMoE over fine-grained MoE.

\begin{table}[h]
    \centering
    \resizebox{\textwidth}{!}{
    \begin{tabular}{cc|ccccccc|c}
        \toprule
        \textbf{Model} & $G$ & \textbf{ARC-c} & \textbf{ARC-e} & \textbf{BoolQ} & \textbf{HellaSwag} & \textbf{LAMBADA} & \textbf{PIQA} & \textbf{RACE} & \textbf{Avg.} \\
        \midrule
        Dense & - & 19.45 & 43.35 & 54.40 & 28.61 & 31.09 & 61.97 & 28.52 & 38.20 \\
        Dense$\times$8 & - & \textbf{22.78} & 48.11 & 59.66 & 31.11 & 35.65 & 65.02 & 29.57 & \textbf{41.70} \\ 
        \midrule
        \multirow{7}{*}{MoE} 
        & 1 & 20.82 & 45.03 & 57.55 & 29.84 & 31.81 & 63.28 & 28.42 & 39.53 \\
        & 2 & 21.42 & 46.55 & 54.25 & 29.95 & 32.52 & 64.09 & 28.61 & 39.62 \\
        & 4 & 20.99 & 46.09 & 55.90 & 30.52 & 35.16 & 63.98 & 29.28 & 40.27 \\
        & 8 & 21.59 & 47.73 & 60.70 & 30.83 & 36.41 & 64.69 & 28.04 & 41.42 \\
        & 16 & 19.80 & 48.82 & 57.34 & 30.64 & 36.00 & 64.74 & 28.71 & 40.86 \\
        & 32 & 21.67 & 48.78 & 57.85 & 31.27 & \textbf{37.10} & 64.69 & 28.52 & 41.41 \\
        & 64 & 20.14 & 48.74 & \textbf{61.50} & 31.03 & 36.31 & 63.93 & 27.85 & 41.35 \\ 
        \midrule
        \multirow{7}{*}{ReMoE} 
        & 1 & 20.22 & 46.68 & 54.16 & 30.26 & 35.94 & 63.55 & 29.38 & 40.03 \\
        & 2 & 20.14 & 47.39 & 57.95 & 30.60 & 34.52 & 63.71 & 28.52 & 40.40 \\
        & 4 & 20.39 & 47.94 & 55.35 & 31.04 & 36.11 & 64.64 & 29.00 & 40.64 \\
        & 8 & 20.82 & 48.36 & 60.49 & 30.90 & 36.06 & 63.87 & 28.90 & 41.34 \\
        & 16 & 21.25 & \textbf{49.41} & 56.06 & 30.91 & 36.23 & 64.91 & 29.95 & 41.25 \\
        & 32 & 20.90 & 48.86 & 55.81 & 31.14 & 36.58 & 64.69 & \textbf{30.05} & 41.15 \\
        & 64 & 20.65 & 48.74 & 60.06 & \textbf{31.56} & 36.43 & \textbf{65.40} & 29.00 & 41.69 \\ 
        \bottomrule
    \end{tabular}}
    \caption{Downstream results of scaling in granularity $G$.}
    \label{tab:scale_g}
\end{table}

\subsection{Load Balancing Ablations}
Table~\ref{tab:lb_or_nolb} presents the downstream evaluation results for the load balancing ablations, as discussed in Section~\ref{sec:w_or_wo_lb}. These results compare performance with and without load balancing, offering insights into the different roles of load balancing in MoE and ReMoE.

\begin{table}[h]
    \centering
    \resizebox{\textwidth}{!}{
    \begin{tabular}{cc|ccccccc|c}
        \toprule
        \textbf{Model} & \textbf{LB} & \textbf{ARC-c} & \textbf{ARC-e} & \textbf{BoolQ} & \textbf{HellaSwag} & \textbf{LAMBADA} & \textbf{PIQA} & \textbf{RACE} & \textbf{Avg.} \\
        \midrule
        Dense & - & 19.45 & 43.35 & 54.40 & 28.61 & 31.09 & 61.97 & 28.52 & 38.20 \\
        MoE & × & 19.20 & 44.74 & 50.80 & 28.60 & 30.18 & 62.24 & 27.94 & 37.67 \\
        MoE & \checkmark & 20.05 & 45.16 & \textbf{57.83} & 29.83 & 32.97 & \textbf{63.55} & 28.33 & 39.67 \\
        ReMoE & × & 19.45 & 46.34 & 56.94 & 30.19 & 31.79 & 63.33 & 28.61 & 39.52 \\
        ReMoE & \checkmark & \textbf{20.22} & \textbf{46.68} & 54.16 & \textbf{30.26} & \textbf{35.94} & \textbf{63.55} & \textbf{29.38} & \textbf{40.03} \\ 
        \bottomrule
    \end{tabular}}
    \caption{Downstream results of training with or without load balancing.}
    \label{tab:lb_or_nolb}
\end{table}

\section{Detailed Results for Domain Specification}
\label{app:domain}

\begin{figure}[p]
    \centering
    \begin{subfigure}{\textwidth}
    \centering
    \includegraphics[width=\linewidth]{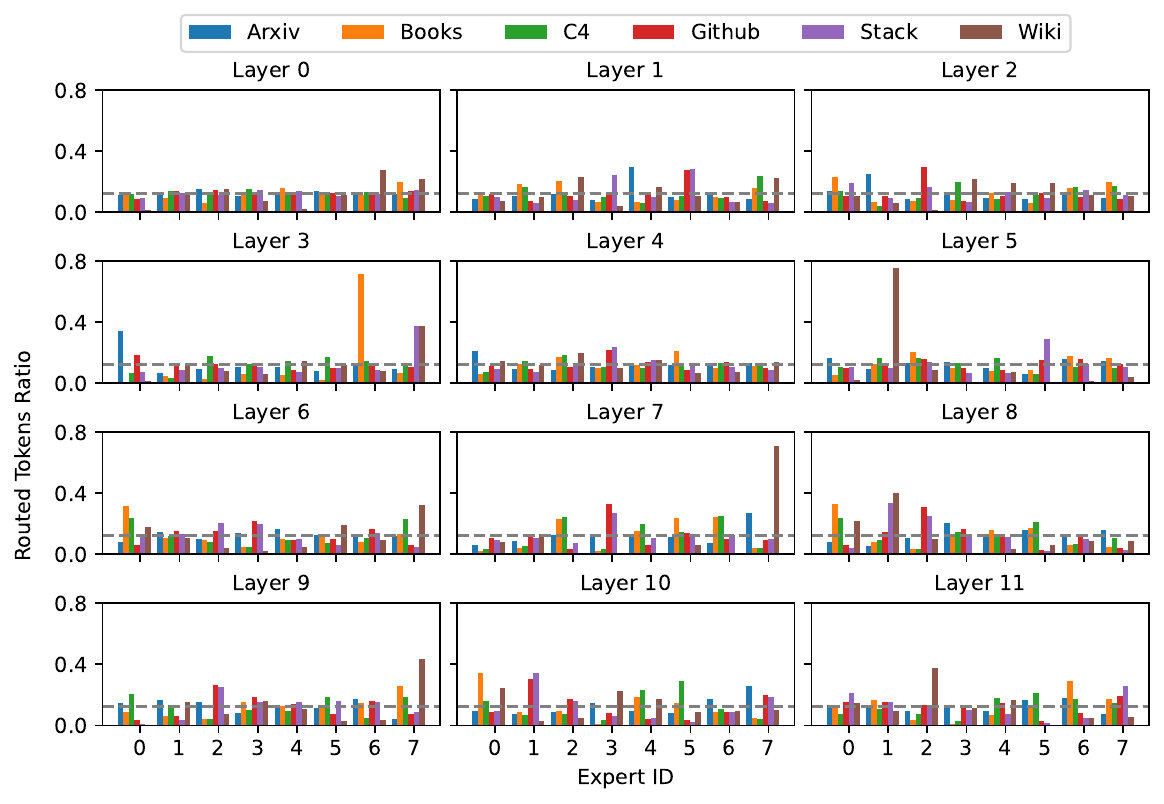}
    \caption{Domain specialization of MoE}
    \label{fig:expert_utilize_all_moe}
        
    \end{subfigure}
    \begin{subfigure}{\textwidth}
    \centering
    \includegraphics[width=\linewidth]{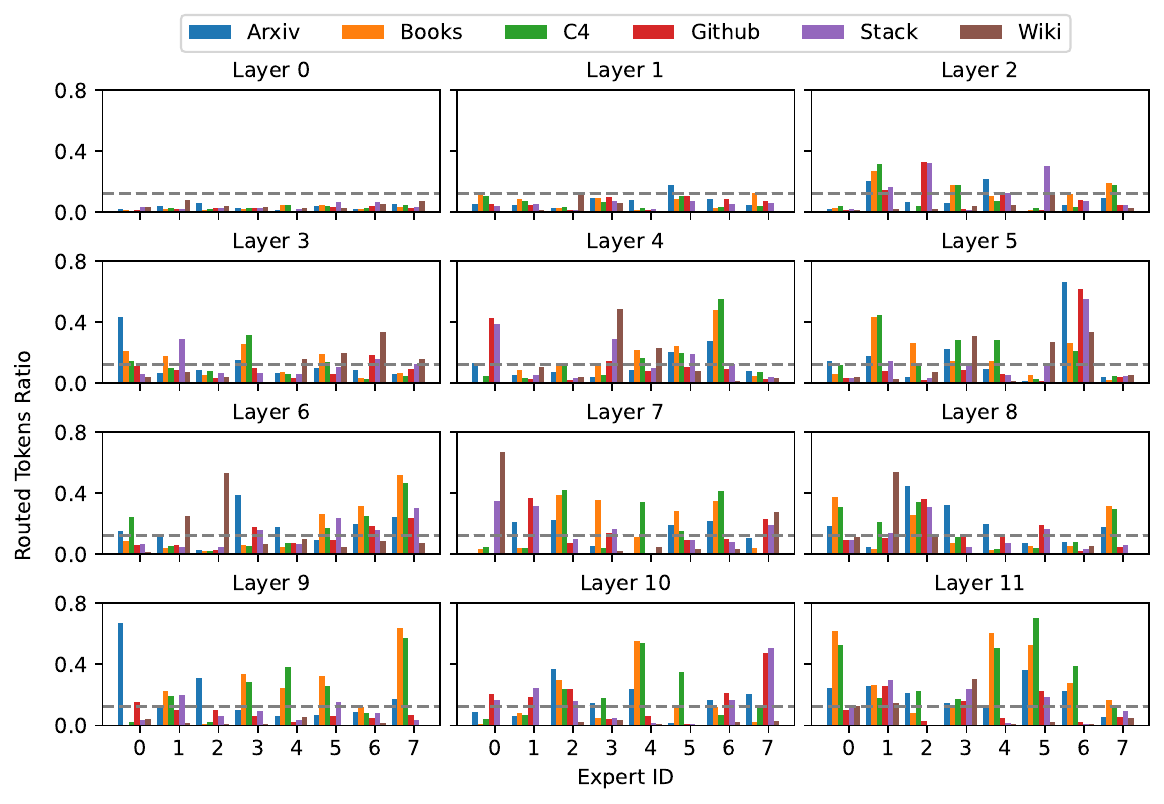}
    \caption{Domain specialization of ReMoE}
    \label{fig:expert_utilize_all}
    \end{subfigure}
    \caption{Detailed results of average routed tokens ratio for MoE and ReMoE in different domains.}
    \label{fig:specilization_all}
\end{figure}

Figure~\ref{fig:specilization_all} shows the average routed tokens ratio of MoE and ReMoE across all layers. ReMoE demonstrates significantly stronger domain specialization compared to MoE, where certain experts are more frequently activated for specific domains. This suggests that ReMoE is better at learning and exploiting the unique characteristics of different domains, allowing it to allocate computational resources more effectively. In contrast, MoE exhibits a more uniform expert activation across domains, indicating less differentiation in its expert specialization.

\newpage

We further analyze the experts in Layer 5 of ReMoE and observe that certain highly related, domain-specific vocabularies are consistently routed to the same expert. To investigate this, we calculate the routing probabilities of different tokens based on their IDs, defined as the ratio of the number of times a specific expert is utilized to the total occurrences of the token. The results are summarized in Table~\ref{tab:route_prob}.

Our findings reveal that the vocabularies exhibit clear specialization, reflecting domain-specific characteristics. For example, Expert 1, which is more frequently assigned to natural language domains (e.g., Books, C4), tends to route tokens such as \texttt{husband}, \texttt{wife}, and \texttt{lover}. In contrast, Expert 6, which is associated with non-natural language domains (e.g., Arxiv, Github, StackExchange), predominantly routes code-related tokens like \texttt{variable}, \texttt{env}, and \texttt{HEAD}.

\begin{table}[h]
    \centering
    \resizebox{\textwidth}{!}{
    \begin{tabularx}{\textwidth}{c|X}
    \toprule
        Expert ID & Routed Tokens With High Probability \\\midrule
        0 & \texttt{End}(100\%);
        \texttt{folding}(100\%); \texttt{Fill}(100\%); \texttt{FILE}(100\%); \texttt{NULL}(100\%);
        \texttt{byte}(100\%);
        \texttt{Release}(99.36\%); \texttt{Del}(99.80\%)\\\midrule
        1 & \texttt{husband}(100\%); \texttt{ife}(100\%); \texttt{baby}(100\%);
        \texttt{human}(100\%); \texttt{lover}(99.60\%); \texttt{).}(99.86\%); \texttt{),}(99.71\%); \texttt{)...}(98.425\%) \\\midrule
        2 & \texttt{invest}(100\%); \texttt{Fortune}(100\%); \texttt{exec} (100\%); \texttt{0000}(100\%); \texttt{Sorry}(100\%); \texttt{bye}(97.82\%); \texttt{If}(97.74\%); \texttt{®}(97.63\%) \\\midrule
        3 & \texttt{Conversely}(100\%); \texttt{Methods}(100\%); \texttt{flower}(100\%); \texttt{Blossom}(99.93\%); \texttt{Argentina}(100\%); \texttt{Georgian}(100\%); \texttt{Uruguay}(98.90\%); \texttt{African} (100\%) \\\midrule
        4 & \texttt{Spring}(100\%); \texttt{Summer}(100\%) \texttt{Autumn}(100\%); \texttt{Winter}(100\%); \texttt{seasons}(99.02\%); \texttt{Temperature} (100\%); \texttt{hot}(97.98\%); \texttt{cold}(100\%)\\\midrule
        5 & \texttt{è}(100\%); \texttt{æ}(99.80\%); \texttt{å}(98.59\%); \texttt{Æ}(97.67\%) \\\midrule
        6 & \texttt{]);}(100\%); \texttt{gif}(100\%); \texttt{size}(100\%); \texttt{variable}(100\%); \texttt{env}(100\%); \texttt{begin}(97.95\%);
        \texttt{HEAD}(97.94\%); \texttt{|}(97.83\%)\\\midrule
        7 & \texttt{Kuala}(100\%); \texttt{Tus}(100\%); \texttt{Lama}(100\%); \texttt{Riley}(98.94\%)\\
        \bottomrule
    \end{tabularx}}
    \caption{Routed tokens with high probability for experts in Layer 5 of ReMoE}
    \label{tab:route_prob}
\end{table}

\section{Domain-Level Dynamic Expert Allocation in ReMoE}
\label{app:dynamic_domain}

\begin{figure}[h]
    \centering
    \includegraphics[width=\linewidth]{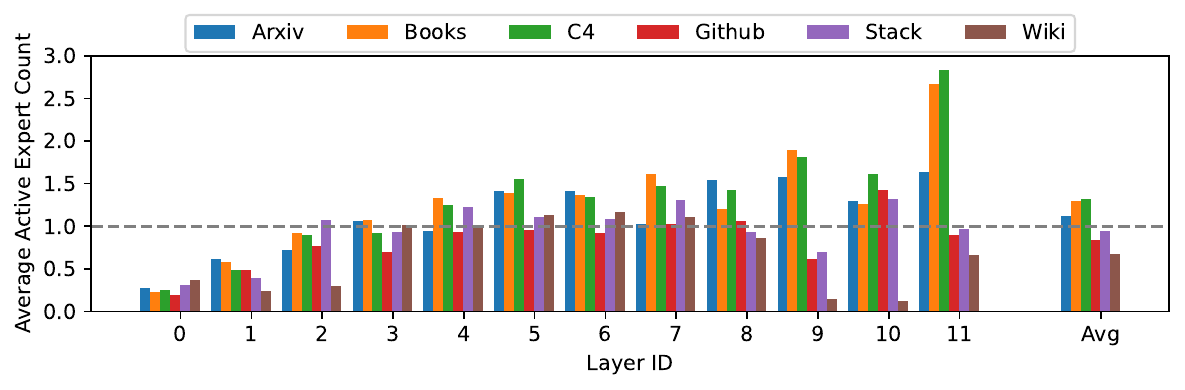}
    \caption{Domain-level dynamic expert allocation}
    \label{fig:domain_sparsity}
\end{figure}

We measure the average active expert count across different domains, as shown in Figure~\ref{fig:domain_sparsity}, and find that the computation allocation in ReMoE also varies at the domain level. Furthermore, this variation increases in deeper layers closer to the output. This is reasonable because deeper layers tend to capture more abstract and domain-specific features, leading to more pronounced specialization in expert activation.

\section{Training MoE with Near-Dense Warmup}

In ReMoE, the training process naturally progresses through three stages, with the first two involving near-dense training where the majority of experts are active. To facilitate a fairer comparison, in Section~\ref{sec:scalability}, we train the MoE model for additional tokens to match the overall computational cost. In this section, we explore an alternative approach by introducing a similar near-dense warmup phase for MoE, referred to as "MoE with warmup," to align its computational footprint with ReMoE across each stage. Specifically, we train the MoE with $N = 182\mathrm{M}$, $E = 8$, and $k = 6$—approximately matching the average sparsity of ReMoE during Stages I and II, as depicted in Figure~\ref{fig:three_stage_sparsity}—for the first 100 steps, before transitioning to $k = 1$ for the remainder of the training process.

Table~\ref{tab:topk_warmup} compares this warmup variant to both standard MoE and ReMoE. The results indicate that the warmup phase provides a modest improvement in validation loss compared to standard MoE, despite matching the overall computational cost. Nonetheless, ReMoE consistently outperforms both variants. This suggests that the three-stage training pipeline learned by ReMoE, with Stages I and II comprising only the first 100 steps, is beneficial to overall performance.

\begin{table}[h]
\centering
\begin{tabularx}{\textwidth}{X|X|XXXXXXX|X}
\toprule
\textbf{Model} & \textbf{Valid Loss} & \textbf{ARC-c} & \textbf{ARC-e} & \textbf{BoolQ} & \textbf{Hella-Swag} & \textbf{LAM-BADA} & \textbf{PIQA} & \textbf{RACE} & \textbf{Avg.} \\ \midrule
MoE             & 1.936                                        & 20.82 & 45.03 & 57.55 & 29.84 & 31.81 & 63.28 & 28.42 & 39.53      \\[0.2cm] 
MoE with warmup & 1.928                                        & 20.73          & 46.38          & 52.35          & 30.28              & 33.90           & 63.76         & 27.66  & 39.29       \\[1cm] 
ReMoE           & 1.921                                      & 20.22          & 46.68          & 54.16          & 30.26              & 35.94           & 63.55         & 29.38  & 40.03         \\ \bottomrule
\end{tabularx}
\caption{Performance of MoE with near-dense warmup}
\label{tab:topk_warmup}
\end{table}

We further extend our experiments with MoE using warmup to configurations with larger $E$, which increases the computational cost of near-dense training. The results, summarized in Table~\ref{tab:warmup_scaling_e}, show that as $E$ increases, the warmup setting consistently improves performance. However, ReMoE still outperforms both variants, maintaining a steeper performance scaling with respect to $E$.

\begin{table}[h]
\centering
\begin{tabularx}{\textwidth}{X|X|X||X|X|X||X|X|X}
\toprule
\textbf{Model, $E=$}8         & \textbf{Valid Loss} & \textbf{Avg. Acc.} & \textbf{Model, $E=$}32         & \textbf{Valid Loss} & \textbf{Avg. Acc.} & \textbf{Model, $E=$}128         & \textbf{Valid Loss} & \textbf{Avg. Acc.} \\ \midrule
MoE                    & 1.936               & 39.53                        & MoE                    & 1.874               & 39.77                        & MoE                    & 1.852               & 41.10                        \\ [0.2cm]
MoE with warmup        & 1.928               & 39.29                        & MoE with warmup        & 1.869               & 40.06                        & MoE with warmup        & 1.841               & 41.34                        \\[1cm] 
ReMoE                  & 1.921               & 40.03                        & ReMoE                  & 1.852               & 41.58                        & ReMoE                  & 1.815               & 42.12                        \\ \bottomrule
\end{tabularx}
\caption{Results for MoE with warmup under different expert count $E$}
\label{tab:warmup_scaling_e}
\end{table}

\begin{figure}
    \begin{minipage}{0.55\linewidth}
        \includegraphics[width=\linewidth]{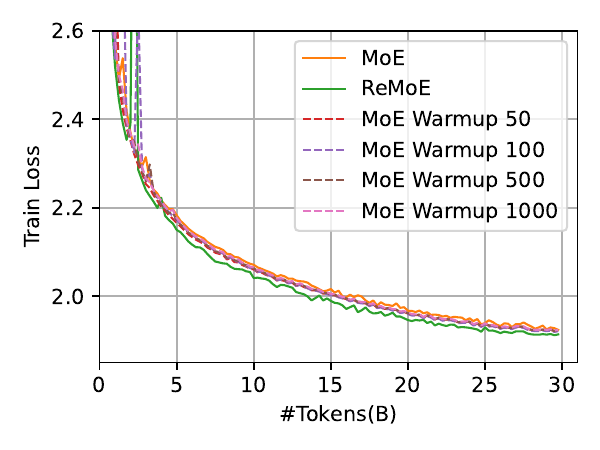}
        \caption{Training curves of MoE with different warmup steps}
        \label{fig:plot_warmup}
    \end{minipage}
    \begin{minipage}{0.38\linewidth}
            \centering
            \begin{tabular}{c|c|c}
            \toprule
                \textbf{Model} & \textbf{Warmup Steps} & \textbf{Valid Loss}\\\midrule
                \multirow{5}{*}{MoE} & 0 & 1.937\\
                 & 50 & 1.930\\
                 & 100 & 1.928\\
                 & 500 & 1.930\\
                 & 1000 & 1.931\\\midrule
                 ReMoE & - & 1.921\\\bottomrule
            \end{tabular}
            \captionof{table}{Final validation loss of MoE with different warmup steps}
            \label{tab:warmup_steps}
    \end{minipage}
    
\end{figure}

To further investigate the impact of warmup steps on MoE performance, we vary the number of warmup steps for the $E=8$ MoE configuration among 50, 100, 500, and 1000. The training curves of these models, along with standard MoE and ReMoE, are shown in Figure~\ref{fig:plot_warmup}, and the final validation losses are summarized in Table~\ref{tab:warmup_steps}.

Our results reveal that performance does not improve monotonically with an increasing number of warmup steps, despite the additional computation. This behavior arises due to the discrepancy between the training objectives of $k=6$ (warmup phase) and $k=1$ (post-warmup phase). For instance, when warmup concludes after 100 steps, the transition between phases is smooth, with the loss changing minimally from $6.491 \rightarrow 6.751$. However, extending warmup to 500 or 1000 steps leads to a more pronounced loss gap of $3.101 \rightarrow 5.827$ and $2.695 \rightarrow 4.428$, respectively.

\newpage
In summary, near-dense warmup can enhance the performance of TopK MoE when training from scratch by providing a better initialization for the experts. However, the warmup phase should conclude while the language model loss is still decreasing rapidly. Prolonging the warmup can exacerbate the gap between the warmup and subsequent training phases, ultimately degrading performance. In contrast, ReMoE naturally determines the appropriate warmup steps and sparsity levels due to its continuous and differentiable training dynamics.

\section{Future Directions}
This work can be advanced in the following ways:
\begin{itemize}
    \item
    \textbf{ReLU Routing for Mixture-of-LoRAs (MoLoRA).}
    MoLoRA \citep{zadouri2023pushing,wu2024mixture,jiao2024rode} integrates MoE architectures to manage multiple Low-Rank Adaptation (LoRA) experts, dynamically activating task-specific adapters during inference. ReMoE’s fully differentiable routing mechanism could enhance MoLoRA by enabling smoother transitions between LoRA experts, particularly when adapters are trained on diverse tasks. Using ReLU straightforwardly in MoLoRA is explored in RoDE \citep{jiao2024rode}, which can be further enhanced by scaling the expert count while controlling the sparsity as in ReMoE.
    \item \textbf{ReLU Routing in Product-Key-Memory (PKM) Networks.} PKM \citep{lample2019large,he_mixture_2024,berges2024memory,huang2024ultra} architectures treat individual neurons as ultra-fine-grained experts, leading to routing complexity at unprecedented scales (e.g., millions of experts). ReMoE’s differentiable routing and steep scaling properties are particularly suited to address PKM’s optimization challenges.
    \item \textbf{Synergy with Efficient Attention Algorithms.} Merging ReMoE’s sparse, conditional feed-forward computation with efficient attention variants—such as quantized~\citep{zhang2024sageattention,zhang2024sageattention2}, linearized~\citep{sun_retentive_2023,gu2023mamba}, sparse~\citep{jiang2025minference,gao2024seerattention}, or mixture-of-attention~\citep{zhang2022mixture,csordas2025switchhead} mechanisms—could enable Transformers to scale efficiently in both sequence length and model capacity without incurring additional computational overhead.  
    \item \textbf{Dynamic Expert Pruning for ReMoE.} ReMoE’s differentiable training inherently promotes expert specialization, with significant variance in expert importance across domains. This property makes ReMoE more amenable to expert pruning~\citep{lu2024not,liu2024efficient} compared to traditional TopK-routed MoE architectures.  
\end{itemize}

%% file: main.bbl
\begin{thebibliography}{64}
\providecommand{\natexlab}[1]{#1}
\providecommand{\url}[1]{\texttt{#1}}
\expandafter\ifx\csname urlstyle\endcsname\relax
  \providecommand{\doi}[1]{doi: #1}\else
  \providecommand{\doi}{doi: \begingroup \urlstyle{rm}\Url}\fi

\bibitem[Ainslie et~al.(2023)Ainslie, Lee-Thorp, de~Jong, Zemlyanskiy, Lebr{\'o}n, and Sanghai]{ainslie2023gqa}
Joshua Ainslie, James Lee-Thorp, Michiel de~Jong, Yury Zemlyanskiy, Federico Lebr{\'o}n, and Sumit Sanghai.
\newblock Gqa: Training generalized multi-query transformer models from multi-head checkpoints.
\newblock \emph{arXiv preprint arXiv:2305.13245}, 2023.

\bibitem[Bengio et~al.(2013)Bengio, L{\'e}onard, and Courville]{bengio2013estimating}
Yoshua Bengio, Nicholas L{\'e}onard, and Aaron Courville.
\newblock Estimating or propagating gradients through stochastic neurons for conditional computation.
\newblock \emph{arXiv preprint arXiv:1308.3432}, 2013.

\bibitem[Berges et~al.(2024)Berges, O{\u{g}}uz, Haziza, Yih, Zettlemoyer, and Gosh]{berges2024memory}
Vincent-Pierre Berges, Barlas O{\u{g}}uz, Daniel Haziza, Wen-tau Yih, Luke Zettlemoyer, and Gargi Gosh.
\newblock Memory layers at scale.
\newblock \emph{arXiv preprint arXiv:2412.09764}, 2024.

\bibitem[Bisk et~al.(2020)Bisk, Zellers, Gao, Choi, et~al.]{bisk2020piqa}
Yonatan Bisk, Rowan Zellers, Jianfeng Gao, Yejin Choi, et~al.
\newblock Piqa: Reasoning about physical commonsense in natural language.
\newblock In \emph{Proceedings of the AAAI conference on artificial intelligence}, volume~34, pp.\  7432--7439, 2020.

\bibitem[Clark et~al.(2022)Clark, de~Las~Casas, Guy, Mensch, Paganini, Hoffmann, Damoc, Hechtman, Cai, Borgeaud, et~al.]{clark2022unified}
Aidan Clark, Diego de~Las~Casas, Aurelia Guy, Arthur Mensch, Michela Paganini, Jordan Hoffmann, Bogdan Damoc, Blake Hechtman, Trevor Cai, Sebastian Borgeaud, et~al.
\newblock Unified scaling laws for routed language models.
\newblock In \emph{International conference on machine learning}, pp.\  4057--4086. PMLR, 2022.

\bibitem[Clark et~al.(2019)Clark, Lee, Chang, Kwiatkowski, Collins, and Toutanova]{clark2019boolq}
Christopher Clark, Kenton Lee, Ming-Wei Chang, Tom Kwiatkowski, Michael Collins, and Kristina Toutanova.
\newblock Boolq: Exploring the surprising difficulty of natural yes/no questions.
\newblock \emph{arXiv preprint arXiv:1905.10044}, 2019.

\bibitem[Clark et~al.(2018)Clark, Cowhey, Etzioni, Khot, Sabharwal, Schoenick, and Tafjord]{clark2018think}
Peter Clark, Isaac Cowhey, Oren Etzioni, Tushar Khot, Ashish Sabharwal, Carissa Schoenick, and Oyvind Tafjord.
\newblock Think you have solved question answering? try arc, the ai2 reasoning challenge.
\newblock \emph{arXiv preprint arXiv:1803.05457}, 2018.

\bibitem[Csord{\'a}s et~al.(2025)Csord{\'a}s, Pi{\'k{e}}kos, Irie, and Schmidhuber]{csordas2025switchhead}
R{\'o}bert Csord{\'a}s, Piotr Pi{\'k{e}}kos, Kazuki Irie, and J{\"u}rgen Schmidhuber.
\newblock Switchhead: Accelerating transformers with mixture-of-experts attention.
\newblock \emph{Advances in Neural Information Processing Systems}, 37:\penalty0 74411--74438, 2025.

\bibitem[Dai et~al.(2024)Dai, Deng, Zhao, Xu, Gao, Chen, Li, Zeng, Yu, Wu, et~al.]{dai2024deepseekmoe}
Damai Dai, Chengqi Deng, Chenggang Zhao, RX~Xu, Huazuo Gao, Deli Chen, Jiashi Li, Wangding Zeng, Xingkai Yu, Y~Wu, et~al.
\newblock Deepseekmoe: Towards ultimate expert specialization in mixture-of-experts language models.
\newblock \emph{arXiv preprint arXiv:2401.06066}, 2024.

\bibitem[Fedus et~al.(2022)Fedus, Zoph, and Shazeer]{fedus2022switch}
William Fedus, Barret Zoph, and Noam Shazeer.
\newblock Switch transformers: Scaling to trillion parameter models with simple and efficient sparsity.
\newblock \emph{Journal of Machine Learning Research}, 23\penalty0 (120):\penalty0 1--39, 2022.

\bibitem[Gale et~al.(2023)Gale, Narayanan, Young, and Zaharia]{gale2023megablocks}
Trevor Gale, Deepak Narayanan, Cliff Young, and Matei Zaharia.
\newblock Megablocks: Efficient sparse training with mixture-of-experts.
\newblock \emph{Proceedings of Machine Learning and Systems}, 5:\penalty0 288--304, 2023.

\bibitem[Gao et~al.(2020)Gao, Biderman, Black, Golding, Hoppe, Foster, Phang, He, Thite, Nabeshima, et~al.]{gao2020pile}
Leo Gao, Stella Biderman, Sid Black, Laurence Golding, Travis Hoppe, Charles Foster, Jason Phang, Horace He, Anish Thite, Noa Nabeshima, et~al.
\newblock The pile: An 800gb dataset of diverse text for language modeling.
\newblock \emph{arXiv preprint arXiv:2101.00027}, 2020.

\bibitem[Gao et~al.(2024)Gao, Zeng, Du, Cao, So, Cao, Yang, and Yang]{gao2024seerattention}
Yizhao Gao, Zhichen Zeng, Dayou Du, Shijie Cao, Hayden Kwok-Hay So, Ting Cao, Fan Yang, and Mao Yang.
\newblock Seerattention: Learning intrinsic sparse attention in your llms.
\newblock \emph{arXiv preprint arXiv:2410.13276}, 2024.

\bibitem[Gu \& Dao(2023)Gu and Dao]{gu2023mamba}
Albert Gu and Tri Dao.
\newblock Mamba: Linear-time sequence modeling with selective state spaces.
\newblock \emph{arXiv preprint arXiv:2312.00752}, 2023.

\bibitem[He(2024)]{he_mixture_2024}
Xu~Owen He.
\newblock Mixture of {A} {Million} {Experts}, July 2024.
\newblock URL \url{http://arxiv.org/abs/2407.04153}.
\newblock arXiv:2407.04153 [cs].

\bibitem[Huang et~al.(2024)Huang, Min, Huang, Zhu, Zeng, Guo, and Zhou]{huang2024ultra}
Zihao Huang, Qiyang Min, Hongzhi Huang, Defa Zhu, Yutao Zeng, Ran Guo, and Xun Zhou.
\newblock Ultra-sparse memory network.
\newblock \emph{arXiv preprint arXiv:2411.12364}, 2024.

\bibitem[Huffman(1952)]{huffman1952method}
David~A Huffman.
\newblock A method for the construction of minimum-redundancy codes.
\newblock \emph{Proceedings of the IRE}, 40\penalty0 (9):\penalty0 1098--1101, 1952.

\bibitem[Jacobs et~al.(1991)Jacobs, Jordan, Nowlan, and Hinton]{jacobs1991adaptive}
Robert~A Jacobs, Michael~I Jordan, Steven~J Nowlan, and Geoffrey~E Hinton.
\newblock Adaptive mixtures of local experts.
\newblock \emph{Neural computation}, 3\penalty0 (1):\penalty0 79--87, 1991.

\bibitem[Jiang et~al.(2024)Jiang, Sablayrolles, Roux, Mensch, Savary, Bamford, Chaplot, Casas, Hanna, Bressand, et~al.]{jiang2024mixtral}
Albert~Q Jiang, Alexandre Sablayrolles, Antoine Roux, Arthur Mensch, Blanche Savary, Chris Bamford, Devendra~Singh Chaplot, Diego de~las Casas, Emma~Bou Hanna, Florian Bressand, et~al.
\newblock Mixtral of experts.
\newblock \emph{arXiv preprint arXiv:2401.04088}, 2024.

\bibitem[Jiang et~al.(2025)Jiang, Li, Zhang, Wu, Luo, Ahn, Han, Abdi, Li, Lin, et~al.]{jiang2025minference}
Huiqiang Jiang, Yucheng Li, Chengruidong Zhang, Qianhui Wu, Xufang Luo, Surin Ahn, Zhenhua Han, Amir Abdi, Dongsheng Li, Chin-Yew Lin, et~al.
\newblock Minference 1.0: Accelerating pre-filling for long-context llms via dynamic sparse attention.
\newblock \emph{Advances in Neural Information Processing Systems}, 37:\penalty0 52481--52515, 2025.

\bibitem[Jiao et~al.(2024)Jiao, Wu, Zhu, Chen, Ngo, and Jiang]{jiao2024rode}
Pengkun Jiao, Xinlan Wu, Bin Zhu, Jingjing Chen, Chong-Wah Ngo, and Yugang Jiang.
\newblock Rode: Linear rectified mixture of diverse experts for food large multi-modal models.
\newblock \emph{arXiv preprint arXiv:2407.12730}, 2024.

\bibitem[Jordan \& Jacobs(1994)Jordan and Jacobs]{jordan1994hierarchical}
Michael~I Jordan and Robert~A Jacobs.
\newblock Hierarchical mixtures of experts and the em algorithm.
\newblock \emph{Neural computation}, 6\penalty0 (2):\penalty0 181--214, 1994.

\bibitem[Kaplan et~al.(2020)Kaplan, McCandlish, Henighan, Brown, Chess, Child, Gray, Radford, Wu, and Amodei]{kaplan2020scaling}
Jared Kaplan, Sam McCandlish, Tom Henighan, Tom~B Brown, Benjamin Chess, Rewon Child, Scott Gray, Alec Radford, Jeffrey Wu, and Dario Amodei.
\newblock Scaling laws for neural language models.
\newblock \emph{arXiv preprint arXiv:2001.08361}, 2020.

\bibitem[Korthikanti et~al.(2023)Korthikanti, Casper, Lym, McAfee, Andersch, Shoeybi, and Catanzaro]{korthikanti2023reducing}
Vijay~Anand Korthikanti, Jared Casper, Sangkug Lym, Lawrence McAfee, Michael Andersch, Mohammad Shoeybi, and Bryan Catanzaro.
\newblock Reducing activation recomputation in large transformer models.
\newblock \emph{Proceedings of Machine Learning and Systems}, 5:\penalty0 341--353, 2023.

\bibitem[Krajewski et~al.(2024)Krajewski, Ludziejewski, Adamczewski, Pi{\'o}ro, Krutul, Antoniak, Ciebiera, Kr{\'o}l, Odrzyg{\'o}{\'z}d{\'z}, Sankowski, et~al.]{krajewski2024scaling}
Jakub Krajewski, Jan Ludziejewski, Kamil Adamczewski, Maciej Pi{\'o}ro, Micha{\l} Krutul, Szymon Antoniak, Kamil Ciebiera, Krystian Kr{\'o}l, Tomasz Odrzyg{\'o}{\'z}d{\'z}, Piotr Sankowski, et~al.
\newblock Scaling laws for fine-grained mixture of experts.
\newblock \emph{arXiv preprint arXiv:2402.07871}, 2024.

\bibitem[Lai et~al.(2017)Lai, Xie, Liu, Yang, and Hovy]{lai2017race}
Guokun Lai, Qizhe Xie, Hanxiao Liu, Yiming Yang, and Eduard Hovy.
\newblock Race: Large-scale reading comprehension dataset from examinations.
\newblock In \emph{Proceedings of the 2017 Conference on Empirical Methods in Natural Language Processing}, pp.\  785--794, 2017.

\bibitem[Lample et~al.(2019)Lample, Sablayrolles, Ranzato, Denoyer, and J{\'e}gou]{lample2019large}
Guillaume Lample, Alexandre Sablayrolles, Marc'Aurelio Ranzato, Ludovic Denoyer, and Herv{\'e} J{\'e}gou.
\newblock Large memory layers with product keys.
\newblock \emph{Advances in Neural Information Processing Systems}, 32, 2019.

\bibitem[Lepikhin et~al.(2020)Lepikhin, Lee, Xu, Chen, Firat, Huang, Krikun, Shazeer, and Chen]{lepikhin2020gshard}
Dmitry Lepikhin, HyoukJoong Lee, Yuanzhong Xu, Dehao Chen, Orhan Firat, Yanping Huang, Maxim Krikun, Noam Shazeer, and Zhifeng Chen.
\newblock Gshard: Scaling giant models with conditional computation and automatic sharding.
\newblock \emph{arXiv preprint arXiv:2006.16668}, 2020.

\bibitem[Lewis et~al.(2021)Lewis, Bhosale, Dettmers, Goyal, and Zettlemoyer]{lewis2021base}
Mike Lewis, Shruti Bhosale, Tim Dettmers, Naman Goyal, and Luke Zettlemoyer.
\newblock Base layers: Simplifying training of large, sparse models.
\newblock In \emph{International Conference on Machine Learning}, pp.\  6265--6274. PMLR, 2021.

\bibitem[Li et~al.(2022)Li, You, Bhojanapalli, Li, Rawat, Reddi, Ye, Chern, Yu, Guo, et~al.]{li2022lazy}
Zonglin Li, Chong You, Srinadh Bhojanapalli, Daliang Li, Ankit~Singh Rawat, Sashank~J Reddi, Ke~Ye, Felix Chern, Felix Yu, Ruiqi Guo, et~al.
\newblock The lazy neuron phenomenon: On emergence of activation sparsity in transformers.
\newblock \emph{arXiv preprint arXiv:2210.06313}, 2022.

\bibitem[Liu et~al.(2024{\natexlab{a}})Liu, Zhu, Lin, Ning, Blaschko, Yan, Dai, Yang, and Wang]{liu2024efficient}
Enshu Liu, Junyi Zhu, Zinan Lin, Xuefei Ning, Matthew~B Blaschko, Shengen Yan, Guohao Dai, Huazhong Yang, and Yu~Wang.
\newblock Efficient expert pruning for sparse mixture-of-experts language models: Enhancing performance and reducing inference costs.
\newblock \emph{arXiv preprint arXiv:2407.00945}, 2024{\natexlab{a}}.

\bibitem[Liu et~al.(2024{\natexlab{b}})Liu, Kim, Wang, Liang, Shen, Cheng, Liu, Tanaka, Wu, Hu, et~al.]{liu2024grin}
Liyuan Liu, Young~Jin Kim, Shuohang Wang, Chen Liang, Yelong Shen, Hao Cheng, Xiaodong Liu, Masahiro Tanaka, Xiaoxia Wu, Wenxiang Hu, et~al.
\newblock Grin: Gradient-informed moe.
\newblock \emph{arXiv preprint arXiv:2409.12136}, 2024{\natexlab{b}}.

\bibitem[Loshchilov(2017)]{loshchilov2017decoupled}
I~Loshchilov.
\newblock Decoupled weight decay regularization.
\newblock \emph{arXiv preprint arXiv:1711.05101}, 2017.

\bibitem[Lu et~al.(2024)Lu, Liu, Xu, Zhou, Huang, Zhang, Yan, and Li]{lu2024not}
Xudong Lu, Qi~Liu, Yuhui Xu, Aojun Zhou, Siyuan Huang, Bo~Zhang, Junchi Yan, and Hongsheng Li.
\newblock Not all experts are equal: Efficient expert pruning and skipping for mixture-of-experts large language models.
\newblock \emph{arXiv preprint arXiv:2402.14800}, 2024.

\bibitem[Muennighoff et~al.(2024)Muennighoff, Soldaini, Groeneveld, Lo, Morrison, Min, Shi, Walsh, Tafjord, Lambert, et~al.]{muennighoff2024olmoe}
Niklas Muennighoff, Luca Soldaini, Dirk Groeneveld, Kyle Lo, Jacob Morrison, Sewon Min, Weijia Shi, Pete Walsh, Oyvind Tafjord, Nathan Lambert, et~al.
\newblock Olmoe: Open mixture-of-experts language models.
\newblock \emph{arXiv preprint arXiv:2409.02060}, 2024.

\bibitem[Muqeeth et~al.(2023)Muqeeth, Liu, and Raffel]{muqeeth2023soft}
Mohammed Muqeeth, Haokun Liu, and Colin Raffel.
\newblock Soft merging of experts with adaptive routing.
\newblock \emph{arXiv preprint arXiv:2306.03745}, 2023.

\bibitem[Narayanan et~al.(2021)Narayanan, Shoeybi, Casper, LeGresley, Patwary, Korthikanti, Vainbrand, Kashinkunti, Bernauer, Catanzaro, et~al.]{narayanan2021efficient}
Deepak Narayanan, Mohammad Shoeybi, Jared Casper, Patrick LeGresley, Mostofa Patwary, Vijay Korthikanti, Dmitri Vainbrand, Prethvi Kashinkunti, Julie Bernauer, Bryan Catanzaro, et~al.
\newblock Efficient large-scale language model training on gpu clusters using megatron-lm.
\newblock In \emph{Proceedings of the International Conference for High Performance Computing, Networking, Storage and Analysis}, pp.\  1--15, 2021.

\bibitem[Paperno et~al.(2016)Paperno, Kruszewski, Lazaridou, Pham, Bernardi, Pezzelle, Baroni, Boleda, and Fern{\'a}ndez]{paperno2016lambada}
Denis Paperno, Germ{\'a}n Kruszewski, Angeliki Lazaridou, Ngoc-Quan Pham, Raffaella Bernardi, Sandro Pezzelle, Marco Baroni, Gemma Boleda, and Raquel Fern{\'a}ndez.
\newblock The lambada dataset: Word prediction requiring a broad discourse context.
\newblock In \emph{Proceedings of the 54th Annual Meeting of the Association for Computational Linguistics (Volume 1: Long Papers)}, pp.\  1525--1534, 2016.

\bibitem[Puigcerver et~al.(2023)Puigcerver, Riquelme, Mustafa, and Houlsby]{puigcerver2023sparse}
Joan Puigcerver, Carlos Riquelme, Basil Mustafa, and Neil Houlsby.
\newblock From sparse to soft mixtures of experts.
\newblock \emph{arXiv preprint arXiv:2308.00951}, 2023.

\bibitem[Rajbhandari et~al.(2020)Rajbhandari, Rasley, Ruwase, and He]{rajbhandari2020zero}
Samyam Rajbhandari, Jeff Rasley, Olatunji Ruwase, and Yuxiong He.
\newblock Zero: Memory optimizations toward training trillion parameter models.
\newblock In \emph{SC20: International Conference for High Performance Computing, Networking, Storage and Analysis}, pp.\  1--16. IEEE, 2020.

\bibitem[Roller et~al.(2021)Roller, Sukhbaatar, Weston, et~al.]{roller2021hash}
Stephen Roller, Sainbayar Sukhbaatar, Jason Weston, et~al.
\newblock Hash layers for large sparse models.
\newblock \emph{Advances in Neural Information Processing Systems}, 34:\penalty0 17555--17566, 2021.

\bibitem[Schulman et~al.(2015)Schulman, Heess, Weber, and Abbeel]{schulman2015gradient}
John Schulman, Nicolas Heess, Theophane Weber, and Pieter Abbeel.
\newblock Gradient estimation using stochastic computation graphs.
\newblock \emph{Advances in neural information processing systems}, 28, 2015.

\bibitem[Sennrich(2015)]{sennrich2015neural}
Rico Sennrich.
\newblock Neural machine translation of rare words with subword units.
\newblock \emph{arXiv preprint arXiv:1508.07909}, 2015.

\bibitem[Shazeer(2020)]{shazeer2020glu}
Noam Shazeer.
\newblock Glu variants improve transformer.
\newblock \emph{arXiv preprint arXiv:2002.05202}, 2020.

\bibitem[Shazeer et~al.(2017)Shazeer, Mirhoseini, Maziarz, Davis, Le, Hinton, and Dean]{shazeer2017outrageously}
Noam Shazeer, Azalia Mirhoseini, Krzysztof Maziarz, Andy Davis, Quoc Le, Geoffrey Hinton, and Jeff Dean.
\newblock Outrageously large neural networks: The sparsely-gated mixture-of-experts layer.
\newblock \emph{arXiv preprint arXiv:1701.06538}, 2017.

\bibitem[Shoeybi et~al.(2019)Shoeybi, Patwary, Puri, LeGresley, Casper, and Catanzaro]{shoeybi2019megatron}
Mohammad Shoeybi, Mostofa Patwary, Raul Puri, Patrick LeGresley, Jared Casper, and Bryan Catanzaro.
\newblock Megatron-lm: Training multi-billion parameter language models using model parallelism.
\newblock \emph{arXiv preprint arXiv:1909.08053}, 2019.

\bibitem[Snowflake(2024)]{snowflake2024arctic}
Snowflake.
\newblock Arctic open: Efficient foundation language models at snowflake, April 2024.
\newblock URL \url{https://www.snowflake.com/blog/arctic-open-efficient-foundation-language-models-snowflake/}.

\bibitem[Song et~al.(2024)Song, Han, Zhang, Hu, Shi, Li, Chen, Liu, Li, Yang, et~al.]{song2024prosparse}
Chenyang Song, Xu~Han, Zhengyan Zhang, Shengding Hu, Xiyu Shi, Kuai Li, Chen Chen, Zhiyuan Liu, Guangli Li, Tao Yang, et~al.
\newblock Prosparse: Introducing and enhancing intrinsic activation sparsity within large language models.
\newblock \emph{arXiv preprint arXiv:2402.13516}, 2024.

\bibitem[Su et~al.(2024)Su, Ahmed, Lu, Pan, Bo, and Liu]{su2024roformer}
Jianlin Su, Murtadha Ahmed, Yu~Lu, Shengfeng Pan, Wen Bo, and Yunfeng Liu.
\newblock Roformer: Enhanced transformer with rotary position embedding.
\newblock \emph{Neurocomputing}, 568:\penalty0 127063, 2024.

\bibitem[Sun et~al.(2023)Sun, Dong, Huang, Ma, Xia, Xue, Wang, and Wei]{sun_retentive_2023}
Yutao Sun, Li~Dong, Shaohan Huang, Shuming Ma, Yuqing Xia, Jilong Xue, Jianyong Wang, and Furu Wei.
\newblock Retentive {Network}: {A} {Successor} to {Transformer} for {Large} {Language} {Models}, August 2023.
\newblock URL \url{http://arxiv.org/abs/2307.08621}.
\newblock arXiv:2307.08621 [cs].

\bibitem[Tan et~al.(2024)Tan, Shen, Panda, and Courville]{tan2024scattered}
Shawn Tan, Yikang Shen, Rameswar Panda, and Aaron Courville.
\newblock Scattered mixture-of-experts implementation.
\newblock \emph{arXiv preprint arXiv:2403.08245}, 2024.

\bibitem[Touvron et~al.(2023)Touvron, Lavril, Izacard, Martinet, Lachaux, Lacroix, Rozi{\`e}re, Goyal, Hambro, Azhar, et~al.]{touvron2023llama}
Hugo Touvron, Thibaut Lavril, Gautier Izacard, Xavier Martinet, Marie-Anne Lachaux, Timoth{\'e}e Lacroix, Baptiste Rozi{\`e}re, Naman Goyal, Eric Hambro, Faisal Azhar, et~al.
\newblock Llama: Open and efficient foundation language models.
\newblock \emph{arXiv preprint arXiv:2302.13971}, 2023.

\bibitem[Vaswani(2017)]{vaswani2017attention}
A~Vaswani.
\newblock Attention is all you need.
\newblock \emph{Advances in Neural Information Processing Systems}, 2017.

\bibitem[Wu et~al.(2024)Wu, Huang, and Wei]{wu2024mixture}
Xun Wu, Shaohan Huang, and Furu Wei.
\newblock Mixture of lora experts.
\newblock \emph{arXiv preprint arXiv:2404.13628}, 2024.

\bibitem[Zadouri et~al.(2023)Zadouri, {\"U}st{\"u}n, Ahmadian, Ermi{\c{s}}, Locatelli, and Hooker]{zadouri2023pushing}
Ted Zadouri, Ahmet {\"U}st{\"u}n, Arash Ahmadian, Beyza Ermi{\c{s}}, Acyr Locatelli, and Sara Hooker.
\newblock Pushing mixture of experts to the limit: Extremely parameter efficient moe for instruction tuning.
\newblock \emph{arXiv preprint arXiv:2309.05444}, 2023.

\bibitem[Zellers et~al.(2019)Zellers, Holtzman, Bisk, Farhadi, and Choi]{zellers2019hellaswag}
Rowan Zellers, Ari Holtzman, Yonatan Bisk, Ali Farhadi, and Yejin Choi.
\newblock Hellaswag: Can a machine really finish your sentence?
\newblock In \emph{Proceedings of the 57th Annual Meeting of the Association for Computational Linguistics}, pp.\  4791--4800, 2019.

\bibitem[Zhang \& Sennrich(2019)Zhang and Sennrich]{zhang2019root}
Biao Zhang and Rico Sennrich.
\newblock Root mean square layer normalization.
\newblock \emph{Advances in Neural Information Processing Systems}, 32, 2019.

\bibitem[Zhang et~al.(2024{\natexlab{a}})Zhang, Huang, Zhang, Wei, Zhu, and Chen]{zhang2024sageattention2}
Jintao Zhang, Haofeng Huang, Pengle Zhang, Jia Wei, Jun Zhu, and Jianfei Chen.
\newblock Sageattention2 technical report: Accurate 4 bit attention for plug-and-play inference acceleration.
\newblock \emph{arXiv preprint arXiv:2411.10958}, 2024{\natexlab{a}}.

\bibitem[Zhang et~al.(2024{\natexlab{b}})Zhang, Huang, Zhang, Zhu, Chen, et~al.]{zhang2024sageattention}
Jintao Zhang, Haofeng Huang, Pengle Zhang, Jun Zhu, Jianfei Chen, et~al.
\newblock Sageattention: Accurate 8-bit attention for plug-and-play inference acceleration.
\newblock \emph{arXiv preprint arXiv:2410.02367}, 2024{\natexlab{b}}.

\bibitem[Zhang et~al.(2022)Zhang, Shen, Huang, Zhou, Rong, and Xiong]{zhang2022mixture}
Xiaofeng Zhang, Yikang Shen, Zeyu Huang, Jie Zhou, Wenge Rong, and Zhang Xiong.
\newblock Mixture of attention heads: Selecting attention heads per token.
\newblock \emph{arXiv preprint arXiv:2210.05144}, 2022.

\bibitem[Zhong et~al.(2024)Zhong, Xia, Chen, and Lewis]{zhong2024lory}
Zexuan Zhong, Mengzhou Xia, Danqi Chen, and Mike Lewis.
\newblock Lory: Fully differentiable mixture-of-experts for autoregressive language model pre-training.
\newblock \emph{arXiv preprint arXiv:2405.03133}, 2024.

\bibitem[Zhou et~al.(2022)Zhou, Lei, Liu, Du, Huang, Zhao, Dai, Le, Laudon, et~al.]{zhou2022mixture}
Yanqi Zhou, Tao Lei, Hanxiao Liu, Nan Du, Yanping Huang, Vincent Zhao, Andrew~M Dai, Quoc~V Le, James Laudon, et~al.
\newblock Mixture-of-experts with expert choice routing.
\newblock \emph{Advances in Neural Information Processing Systems}, 35:\penalty0 7103--7114, 2022.

\bibitem[Zoph et~al.(2022)Zoph, Bello, Kumar, Du, Huang, Dean, Shazeer, and Fedus]{zoph2022st}
Barret Zoph, Irwan Bello, Sameer Kumar, Nan Du, Yanping Huang, Jeff Dean, Noam Shazeer, and William Fedus.
\newblock St-moe: Designing stable and transferable sparse expert models.
\newblock \emph{arXiv preprint arXiv:2202.08906}, 2022.

\bibitem[Zuo et~al.(2021)Zuo, Liu, Jiao, Kim, Hassan, Zhang, Zhao, and Gao]{zuo2021taming}
Simiao Zuo, Xiaodong Liu, Jian Jiao, Young~Jin Kim, Hany Hassan, Ruofei Zhang, Tuo Zhao, and Jianfeng Gao.
\newblock Taming sparsely activated transformer with stochastic experts.
\newblock \emph{arXiv preprint arXiv:2110.04260}, 2021.

\end{thebibliography}
